\documentclass[10pt,letterpaper]{article}
\usepackage[margin=1in]{geometry}
\usepackage[utf8]{inputenc}
\usepackage[T1]{fontenc}
\usepackage[english]{babel}
\usepackage[onehalfspacing]{setspace}

\usepackage[numbers]{natbib}
\usepackage{graphicx} 
\usepackage{float}
\usepackage{algorithm}  
\usepackage{algpseudocode}
\usepackage{color}
\usepackage{setspace}
\usepackage{diagbox}
\usepackage{caption}
\usepackage{booktabs}
\usepackage{amsmath}
\usepackage{siunitx}
\usepackage{textcomp}
\usepackage{multirow}
\usepackage{makecell}
\usepackage{url}

\newcommand{\unet}{U-Net\xspace}

\newcommand\blue[1]{\textcolor{black}{#1}}
\newcommand\black[1]{\textcolor{black}{#1}}

\newcommand{\norain}{\textsc{Others}\xspace}
\newcommand{\lightrain}{\textsc{Light}\xspace}
\newcommand{\heavyrain}{\textsc{Heavy}\xspace}
\newcommand{\lightheavyrain}{\textsc{Rain}\xspace}

\usepackage{subcaption}
\usepackage[shortlabels]{enumitem}
\usepackage{xspace}
\usepackage{amssymb}
\usepackage[dvipsnames]{xcolor}
\usepackage{epstopdf}

\begin{document}
\title{Effective Training Strategies for Deep-learning-based\\ Precipitation Nowcasting and Estimation}
\author{Jihoon Ko\,$^{1,*}$, Kyuhan Lee\,$^{1,*}$, Hyunjin Hwang\,$^{1,*}$,\\Seok-Geun Oh\,$^{2}$, Seok-Woo Son\,$^{2}$, and Kijung Shin\,$^{1}$}
\date{\normalsize $^{1}$ KAIST, Seoul, South Korea, \\ 
	$^{2}$ Seoul National University, Seoul, South Korea, \\
	jihoonko@kaist.ac.kr, kyuhan.lee@kaist.ac.kr, hyunjinhwang@kaist.ac.kr,\\seokgeunoh@snu.ac.kr, seokwooson@snu.ac.kr, kijungs@kaist.ac.kr
}
\maketitle
\def\thefootnote{*}\footnotetext{Equal contribution.}
\def\thefootnote{\arabic{footnote}}

\begin{abstract}
Deep learning has been successfully applied to precipitation nowcasting. In this work, we propose a pre-training scheme and a new loss function for improving deep-learning-based nowcasting. First, we adapt U-Net, a widely-used deep-learning model, for the two problems of interest here: precipitation nowcasting and precipitation estimation from radar images.
We formulate the former as a classification problem with three precipitation intervals and the latter as a regression problem.
For these tasks, we propose to pre-train the model to predict radar images in the near future without requiring ground-truth precipitation, and we also propose the use of a new loss function for fine-tuning to mitigate the class imbalance problem.
We demonstrate the effectiveness of our approach using radar images and precipitation datasets collected from South Korea over seven years.
It is highlighted that our pre-training scheme and new loss function improve the \blue{critical success index (CSI)} of nowcasting of heavy rainfall (at least $10$ mm/hr) by up to $\mathbf{95.7\%}$ and $\mathbf{43.6\%}$, respectively, at a 5-hr lead time.
We also demonstrate that our approach reduces the precipitation estimation error by up to $\mathbf{10.7\%}$, compared to the conventional approach, for light rainfall (between $1$ and $10$ mm/hr). 
Lastly, we report the sensitivity of our approach to different resolutions and a detailed analysis of four cases of heavy rainfall. 
\end{abstract}

\section{Introduction}
\label{sec:intro}
Modern weather forecasts are largely based on Numerical Weather Prediction (NWP) models that describe the essential physical processes in the atmosphere and ocean, at the surface, and in the soil using the fluid dynamic and thermodynamic laws \citep{shuman1989history,harper200750th, marchuk2012numerical}. Numerous studies have demonstrated that NWP models can reasonably resolve the convective development-decay and realistically conduct precipitation forecasts  \citep{benjamin2004hourly,shrestha2013evaluation,sun2014use,wang2016quantitative}. \blue{However, the ability of NWP models to undertake precipitation nowcasting, which generally considers the lead time of 1-6 h \citep{agrawal2019machine, sonderby2020metnet, ravuri2021skillful}, still faces challenges. For example, \citet{shrestha2013evaluation} observed that the NWP model underestimates or overestimates precipitation nowcasting in southeastern Australia depending on the spatial and temporal model resolutions. \cite{sun2014use} noted the spinup issue, which degrades the performance of the model in first few hours due to poor initialization. This issue is substantially overcome with recent techniques such as the rapid update cycle and sophisticated data assimilation, with the requirements, however, of high-quality observations and enormous computational resources \citep{benjamin2004hourly, sun2014use}.}

Another popular traditional approach for precipitation forecasting is the optical flow method. Optical flow refers to a method by which to calculate the motions of image intensities, and it has been used widely for precipitation nowcasting based on radar reflectivity.
Specifically, it extrapolates radar reflectivity by estimating the velocity fields. 
Despite its lower computational complexity, optical flow often shows better predictive performance than NWP for precipitation nowcasting \citep{bechini2017enhanced, bowler2004development}.
Moreover, optical flow has been combined with NWP to realize the advantages of both approaches \citep{seed2013formulation, hwang2015improved}. 
However, it is difficult to determine the model parameters, especially because the two steps of the optical flow (i.e., velocity field estimation and radar extrapolation) are performed separately.

With the rise of deep learning in recent decades, deep learning has been applied in many fields, and precipitation nowcasting is one of them.
Training deep-learning models requires archived data and substantial computations, and changes in data characteristics (e.g. updating radars) may require new archived data for training.
However, once they are trained, making inferences using them requires much less computation.
This feature allows newly arriving data to be incorporated quickly into a prediction, making it particularly advantageous for precipitation nowcasting.

For the problem of precipitation nowcasting from radar images,
\cite{shi2015convolutional} formulated precipitation nowcasting from radar echo images as a spatiotemporal sequence forecasting problem, and for this problem, they proposed a convolutional LSTM (ConvLSTM).
\cite{shi2017deep} further improved ConvLSTM by additionally learning a location-variant structure for recurrent connections in the model, and
\cite{sonderby2020metnet} improved ConvLSTM by adding a spatial aggregator based on axial self-attention \citep{ho2019axial} at the end of the model, instead of directly modifying the architecture of ConvLSTM.
\cite{agrawal2019machine} proposed the use of U-Net \citep{ronneberger2015u}, which proved highly effective for various computer vision tasks, for nowcasting. Their method outperformed a strong optical flow-based method \citep{pulkkinen2019pysteps} and HRRR \citep{benjamin2016north} on one hour nowcasting, though, it was consistently outperformed by HRRR when the prediction window exceeded five hours. \cite{lebedev2019precipitation} combined a U-Net based model with optical flow techniques for precipitation nowcasting from radar and satellite images, and \cite{ayzel2020rainnet} used a U-Net-based model to predict precipitation after five minutes and proposed to feed its output back into the model for longer-term predictions.

In order to realize the capacity of deep neural networks fully, it is known that not only the model architectures but also training methods used are important. For example, to apply deep-learning models to precipitation nowcasting, the class imbalance problem should be considered \citep{japkowicz2002class}. That is, if the number of heavy rainfall cases is relatively small as compared to that of non-precipitation cases and light rainfall cases, classifiers are easily biased toward majority classes with many examples such that deep-learning models show poor predictive performance for heavy rainfall, which must be precisely predicted. \cite{lin2017focal} mitigated this problem in dense object detection with highly imbalanced data,  proposing the \textit{focal loss}, which allows the models being trained rapidly to focus on challenging examples by down-weighting the importance of easy examples during training. 
In order to achieve a higher F1 score, which is widely used for evaluating classifiers when classes are imbalanced, \cite{lebedev2019precipitation} proposed the use of the \textit{dice loss} \citep{sudre2017generalised} in addition to binary cross entropy, and \cite{kaggle} designed a differentiable loss function close to the F1 score, empirically demonstrating that it is highly correlated with the F1 score.

Pre-training deep-learning models recently becomes a dominant paradigm in computer vision and image processing \citep{donahue2014decaf,Girshick_2014_CVPR,doersch2017multi,xie2018pre}.
Pre-training a deep-learning model refers to training the model for related tasks, which are typically supervised with larger data or unsupervised without requiring scarce labels, to help the model learn parameters useful for target tasks. After being pre-trained, the model is fine-tuned for the target tasks.
For example, \cite{Girshick_2014_CVPR} improved the accuracy of convolutional neural networks (CNNs) for object detection by pre-training them on large image datasets from different domains. \cite{doersch2017multi} pre-trained CNNs for multiple self-supervised tasks without manual labeling, demonstrating improved performances on image classification, object detection, and depth prediction.
Regarding weather-related applications, \cite{rasp2021data} proposed to pre-train ResNet \citep{he2016deep} using historical climate model outputs (e.g., CMIP6 data \citep{eyring2016overview}) before fine-tuning via reanalysis data (e.g., ERA5 data \citep{hersbach2020era5}) for medium-range weather forecasting from six hours to five days ahead. Their empirical analysis showed that using climate model outputs for pre-training prevents overfitting and thus leads to performance gains in prediction tasks, except for precipitation estimations. 
It should be noted that their approach is not suitable for precipitation nowcasting given that it is based on climate model outputs and reanalysis data, which are scarcely obtained in real time. Moreover, predictions in this case are a much lower spatial resolution than the resolutions considered in this work.

Motivated by the success of deep learning in precipitation nowcasting, we propose effective training strategies for improving such deep-learning-based approaches, especially those based on radar reflectivity images.
We propose to pre-train deep-learning models for predicting radar reflectivity before fine-tuning them for precipitation nowcasting based on two observations. 
While deep-learning models commonly require a vast amount of data for effective training, the availability of ground-truth precipitation is restricted to weather stations. On the other hand, radar reflectivity is typically available for a wide range of regions, making it easy to obtain large amounts of data for effectively training deep-learning models. 
Given that radar reflectivity is generally correlated with precipitation rates, the parameters of deep-learning models trained for radar reflectivity are likely to be useful for nowcasting. We also propose a novel loss function for mitigating the class imbalance problem, which is commonly found in precipitation data.
Decreasing our loss function requires deep-learning models to improve the predictive performance not only for majority classes but also minority classes.
Thus, training deep-learning models aiming to minimize our loss function prevents them from being biased toward majority classes.

This study introduces a new deep-learning model in conjuction with effective training strategies and a novel loss function and evaluates the capability of the model in performance precipitation forecast with up to six hours of lead time. Radar and in-situ precipitation datasets across South Korea for the period of 2014-2020 are used during the model training and evaluation processes.
In Section~\ref{sec:problemdef}, first we define the two tasks, i.e., precipitation nowcasting and precipitation estimation, and the notations used.
Next, we describe our concrete methods for the model. The model architecture, pre-training methodology, and details of the loss functions are also described in this section.
In Section~\ref{sec:experiments}, we show the experimental results of the two tasks and analyze the performance outcomes.
\blue{In Section~\ref{sec:discussion}, we compare our work with recent studies of deep learning-based nowcasting. Lastly, we conclude our work in Section~\ref{sec:conclusion}.}

\section{Methodology}
\label{sec:problemdef}
\label{sec:method}
In this section, we formulate the two problems of interest: precipitation nowcasting and precipitation estimation. Then, we describe our deep-learning model for these two problems. Lastly, we describe the training methods for the model, with a focus on our proposed pre-training scheme and loss function. Table~\ref{tab:notations} lists the notations frequently used throughout the paper. 

\begin{table}[t]
\centering
\caption{Frequently-used notations.}\label{tab:notations}
\scalebox{0.9}{
    \begin{tabular}{r|l}
\toprule
\textbf{Notation} & \textbf{Description} \\
\midrule
$R_t$ & Radar (reflectivity) image at time $t$ \\
$I$ & Set of regions of interest (i.e., a subset of pixels in radar images) \\ 
$I'$ & Subset of $I$ where weather stations exist \\
$i$ & Region index \\
\midrule
$R_{t^*,i}$ & Ground-truth radar reflectivity at time $t^{*}$ in region $i$ \\
$\hat{R}_{t^*,i} $ & Predicted probability distribution over radar reflectivity classes at time $t^*$ in region $i$ \\
$\hat{R}_{t^*,i,r} $ & Predicted probability for the radar reflectivity class $r$ at time $t^*$ in region $i$ \\
\midrule
$P_{t^*,i}$  & Ground-truth precipitation class at time $t^*$ in region $i$ \\
$\hat{P}_{t^*, i} $ & Predicted probability distribution over precipitation classes at time $t^*$ in region $i$ \\
$\hat{P}_{t^*, i,p} $ & Predicted probability for the precipitation class $p$ at time $t^*$ in region $i$ \\
\midrule
$C_{t, i}$ &  Ground-truth precipitation accumulated from time $t-60$ to $t$ in region $i$\\
$\hat{C}_{t, i}$ &  Estimated accumulative precipitation from time $t-60$ to $t$ in region $i$\\
\bottomrule
\end{tabular}
}
\end{table}

\subsection{Problem definition}

\subsubsection{Precipitation nowcasting}
\label{sec:problemdef:nowcasting}

The objective of \textit{precipitation nowcasting} is to predict precipitation levels in the near future.
In our setting, we are given seven radar (reflectivity) images. Each pixel in the images indicates the radar reflectivity in units of dBZ at the corresponding region\footnote{dBZ stands for decibel relative to $Z$. It is a logarithmic unit for comparing the equivalent radar reflectivity factor ($Z$) of  remote objects. 
The radar reflectivity factor $Z$ corresponds to $10 \cdot \log_{10} Z/Z_{0}$ dbZ, where $Z_{0}$ is the return of a water droplet with a diameter of 1 mm in a meter cube.} 
for the past hour \black{in ten-}minute interval\black{s}. Using the given images, we aim to predict the precipitation at each region of interest up to six hours.
Specifically, if we let $t$ be the current time in minutes, we aim to predict precipitation accumulated from time $t^{*}-60$ to $t^{*}$, for each $t^{*}\in \{t+60, t+120, \cdots, t+360\}$, at each region (i.e., each pixel in radar images) of interest.
We formulate the problem as a classification problem as in previous studies \citep{agrawal2019machine,sonderby2020metnet},\footnote{\blue{As a result, our deep-learning model produces a probabilistic distribution over classes (e.g., a 30\% chance of heavy \heavyrain, a 30\% chance of \lightrain, and a 40\% chance of \norain). If the nowcasting problem is formulated as a regression task, while deep-learning models can produce a specific point estimate, it can be difficult to distinguish a 10\% chance of 30mm/hr from a 100\% chance of 3mm/hr, as their expected precipitation rate is the same.}} and to this end, we categorize precipitation into three classes: \heavyrain (at least $10$ mm/hr), \lightrain (between $1$ and $10$ mm/hr), and \norain (less than $1$ mm/hr).
Our tasks are mainly focused on \lightrain, \heavyrain, and \lightheavyrain $(=\lightrain+ \heavyrain)$.
We denote \black{the} seven \black{given} radar images \black{as} $R_{t-60}, R_{t-50}, \cdots, R_t$, and we use $P_{t^{*},i}$ to indicate the class of precipitation between time $t^{*}-60$ and $t^{*}$ \black{in the} given region $i$.
Then, the precipitation nowcasting problem is formulated as learning a function $\Phi$,
which predicts the class ${P}_{t^{*},i}$ \black{in} each region $i$  as follows:
\begin{equation}
    \{(i,\hat{P}_{t^{*},i})\}_{i\in I}=\Phi(R_{t-60}, R_{t-50}, …, R_t, t^{*}),
\end{equation}
where $I$ denotes the set of regions of interest \black{(i.e., a subset of pixels in radar images)}, and $\hat{P}_{t^{*},i}$ denotes the probability \black{distribution over} the three classes.

\subsubsection{Precipitation estimation}
\label{sec:problemdef:estimation}
The objective of \textit{precipitation estimation} is to estimate precipitation accumulated over time from a series of radar images.
In our setting, given seven radar images  $R_{t-60}, R_{t-50}, \cdots, R_{t}$ with ten-minute intervals, we aim to estimate the precipitation $C_{t,i}$ between time $t-60$ to $t$ (in minutes) \black{in} each region $i$.
This problem is formulated as a regression problem, and we aim to learn a function $\Psi$, which estimates $C_{t,i}$ \black{in} each region $i$ as follows:
\begin{equation}
\{(i,\hat{C}_{t,i})\}_{i\in I} = \Psi(R_{t-60}, R_{t-50}, …, R_t), \label{eq:function:estimation}
\end{equation}
where $I$ denotes the set of regions, and $\hat{C}_{t,i}$ denotes \black{the} estimate of $C_{t,i}$.

\subsection{Deep-learning model}
\label{sec:method:model}

\begin{figure}[t]
    \centering
    \includegraphics[width=\linewidth]{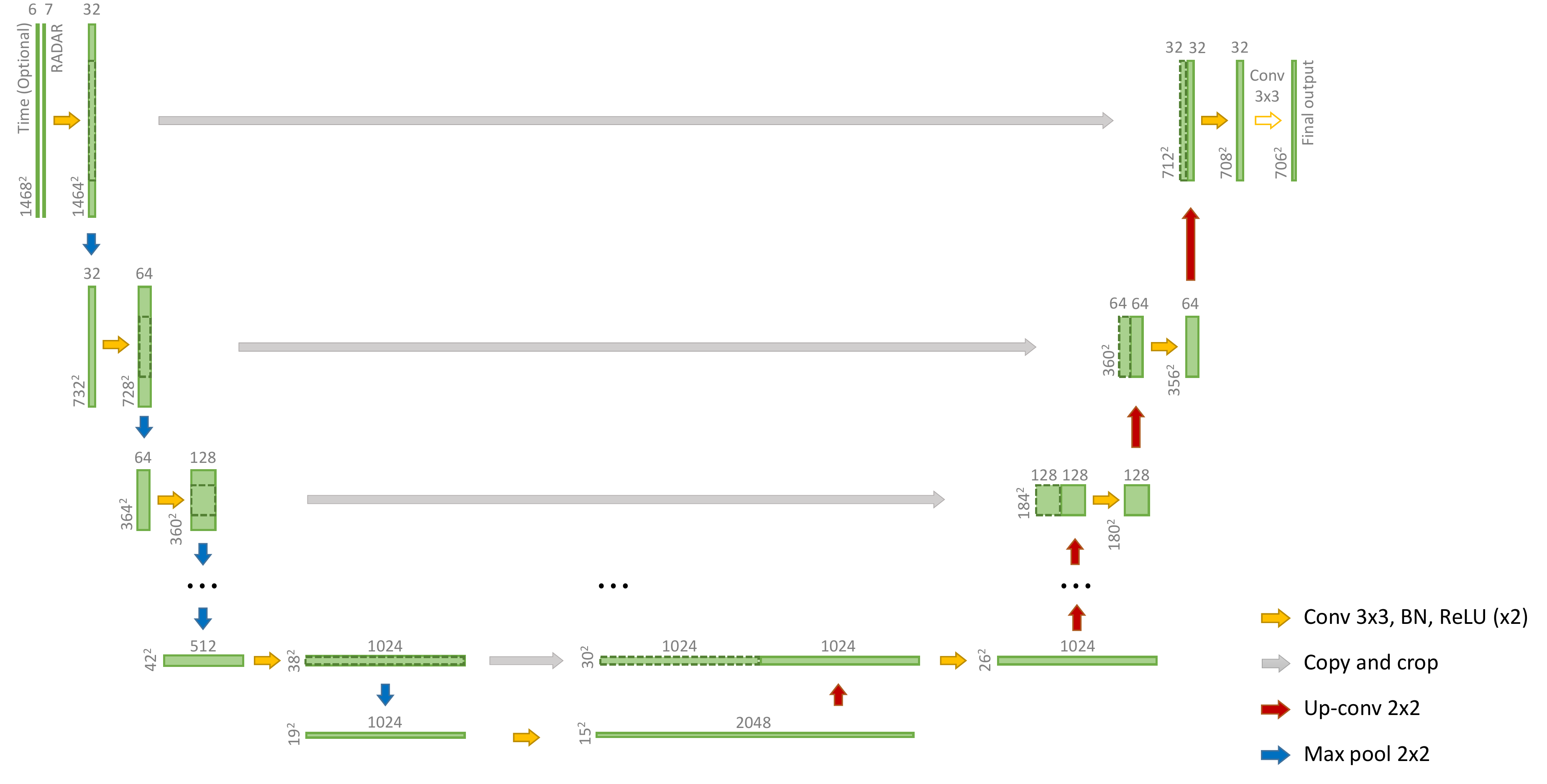}
    \caption{\black{Pictorial description} of our deep-learning model, where U-Net \citep{ronneberger2015u} is tailored for the problems of precipitation nowcasting and estimation.
    Our model consists of seven steps of downsampling (depicted \black{by the} blue arrows) and seven steps of upsampling (depicted \black{by the} red arrows) to make the width and height of the coarsest feature map roughly $2^4=16$. 
    In our setting (see Sections~\ref{sec:problemdef} and \ref{sec:experiments:settings}), the input has seven channels, which correspond to seven consecutive radar images, and for precipitation nowcasting,  six \black{additional} channels for encoding the target time.
    At $1$ km $\times$ $1$ km resolution, the input dimensions are $1468\times 1468$, \black{and} the output dimensions are $706\times 706$.
    \black{That is, the set of regions of interest is \black{a} patch of \black{a} dimension \black{of} $706\times 706$ at the center of radar images.}
    \black{A detailed description of our deep-learning model is provided in Appendix~\ref{sec:app:model}.}
    }
    \label{fig:model}
\end{figure}

In this work, we extend a \unet based model \citep{agrawal2019machine} for our problems. Despite the success of ConvLSTM and its variants, our preliminary study demonstrated that, in our experimental settings (see Section~\ref{sec:experiments:settings}), U-Net based models show similar or higher accuracy than ConvLSTM-based models.
Thus, in the rest of this paper, we focus on a U-Net based model, while our training schemes are not specialized to any specific model.

\subsubsection{U-Net}
\unet \citep{ronneberger2015u} is a CNN devised originally for image segmentation tasks, where the objective is to link each pixel of an image to a corresponding class.
\unet consists of a contracting path and an expansive path, forming a U-shaped architecture. In the contracting path (encoder), spatial information is reduced through repeated convolution operations without padding, and the number of feature channels is doubled to capture the high-resolution context. In the expansive path (decoder), the high-resolution spatial information decoded from up convolution and the feature information obtained from the contracting path are concatenated, making it possible to assemble a precise output and thus enhance the image segmentation tasks.

\unet has been applied to a variety of problems, and notably it also proved successful for precipitation nowcasting \citep{agrawal2019machine}.
In \black{that} work, each pixel in a series of radar images is assigned to a class according to the instantaneous rate of precipitation at the corresponding location after a fixed amount of time, and a U-Net-based model is used to predict the class of each pixel.

\subsubsection{Our extension}
A pictorial description of our model is given in Figure~\ref{fig:model}\black{, and we described the details in Appendix~\ref{sec:app:model}.}
For the precipitation nowcasting problem (see Section~\ref{sec:problemdef:nowcasting}), we extend a U-Net based model for precipitation nowcasting \citep{agrawal2019machine}
so that it can predict the precipitation at any given target time $t^*\in \{t+60, t+120, \cdots, t+360\}$, where $t$ is the current time in minutes\black{.}
The target time is encoded as a one-hot vector for each pixel, which is the \black{identical} for every pixel, and the vectors together form a tensor.
The tensor is concatenated to the input radar images, 
\black{with} the number of input channels \black{thus equaling} the sum of the number of the input radar images (i.e., seven) and the number of possible target times (i.e., six). 
Note that the target time is not required in the precipitation estimation problem (see Section~\ref{sec:problemdef:estimation})\black{; therefore,} only the radar images are used as the input of our model.

\subsection{Training strategies}
\label{sec:method:training}

\begin{figure}[t]
    \centering
    \includegraphics[width=\linewidth]{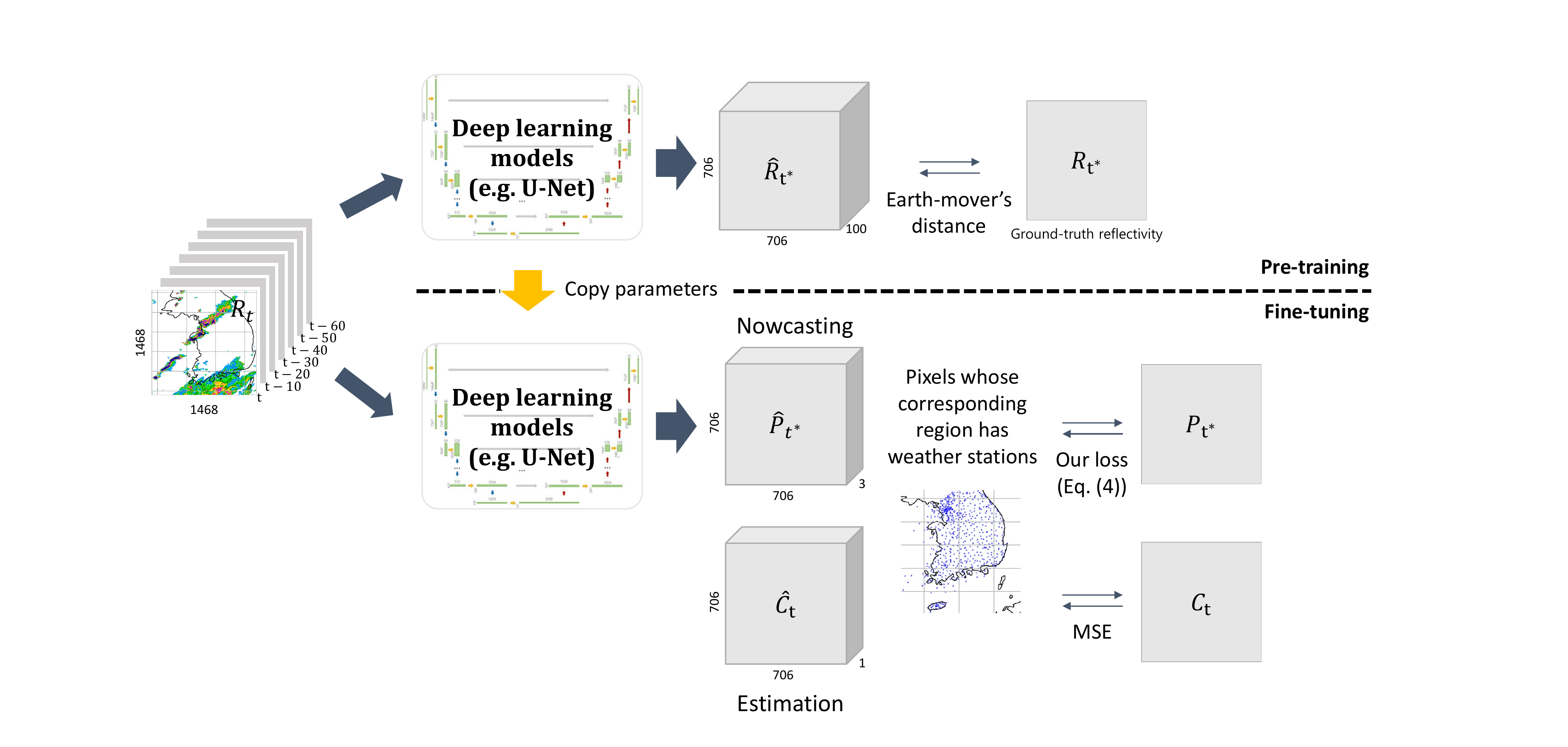}
    \caption{Proposed training procedure consisting of the pre-training phase (top) and the fine-tuning phase (bottom).
    In the pre-training phase (see Section~\ref{sec:method:training:pre}), we train a U-Net based deep-learning model for predicting radar reflectivity in the near future.
    In the fine-tuning phase (see Section~\ref{sec:method:training:tune}), we train the deep-learning model initialized with the pre-trained parameters for one of the target tasks described in Section~\ref{sec:problemdef}.
    When the target task is precipitation nowcasting, our novel loss function in Eq.~\eqref{eq:loss:nowcasting} is used \black{to} mitigat\black{e} the class imbalance problem.
    }
    \label{fig:training_details}
\end{figure}

We describe our proposed training schemes, which consist of two phases: (a) \textbf{pre-training phase}: training a deep-learning model for predicting radar reflectivity in the near future, \black{and} (b) \textbf{fine-tuning phase}: fine tuning the deep-learning model initialized with the pre-trained parameters for each of the target tasks described in Section~\ref{sec:problemdef}.
The overall training process is described in Figure~\ref{fig:training_details}\black{.}

\subsubsection{Pre-training phase}
\label{sec:method:training:pre}
The objective of the pre-training phase is to meaningfully initialize the parameters of deep-learning models, which are later fine-tuned for \black{the }target tasks\black{.}
We consider the problem of predicting the radar image $R_{t^{*}}$ at each target time $t^*\in \{t+60, t+120, \cdots, t+360\}$, where $t$ is the current time in minutes.
Specifically, we predict the radar reflectivity in units of dBZ at each pixel of interest by modeling \black{the probability distribution over $r_{\max}$ classes}, where $r_{\max}$ is the maximum radar reflectivity that we consider.\footnote{In our experiments, we set $r_{\max}$ to $100$ and clamped each input \black{reflectivity value} so that \black{these values} lie between $-0.5$ to $r_{\max} - 0.5$. \black{Specifically, we replaced each actual reflectivity $r$ with $\min(\max(-0.5, r), r_{\max} - 0.5)$.}}
We use $\hat{R}_{t^{*},i}$ to denote the output after softmax activation\footnote{\black{When the raw output $\tilde{R}_{t^{*},i,r}$ of the model is given for each $r$, the output $\hat{R}_{t^{*},i,r}$ after the activation is defined as $\exp(\tilde{R}_{t^{*},i,r}) / \sum_{r=1}^{r_{\max}} \exp(\tilde{R}_{t^{*},i,r})$.}} at each pixel $i$, \black{with} \black{$\hat{R}_{t^{*},i,r}$ \black{denoting} the predicted probability of the reflectivity lying between $(r-1.5)$ and $(r-0.5)$ dBZ.}
We use the earth-mover\black{'s} distance \citep{rubner2000earth} between $\hat{R}_{t^{*}}$ and $R_{t^{*}}$ as the loss function as follows:
\begin{equation}
    L_{pre-training} = \sum_{R_{t}\in D} \sum_{t^*\in T_t} \sum_{i \in I} \sum_{r=1}^{r_{\max}} \left(\hat{R}_{t^{*}, i, r} |(r - 1) -  R_{t^{*},i}|\right), \label{eq:loss:pretrain}
\end{equation}
where $D$ is the set of radar images in the training set and $T_t$ is the set of potential target times. In the pre-training phase for precipitation nowcasting, $T_{t}=\{t+60, t+120, \cdots, t+360\}$. In the pre-training phase for precipitation estimation, $T_{t}=\{t\}$.
Note that $(r-1)$ in Eq.~\eqref{eq:loss:pretrain} is the center of the interval between $(r-1.5)$ and $(r-0.5)$, and the probability of $R_{t^{*},i}$ lying in the interval is predicted to be $\hat{R}_{t^{*}, i,r}$.

We choose the problem of predicting radar reflectivity for pre-training for two reasons. First, it is known that radar reflectivity and precipitation rates are \black{generally} correlated \citep{marshall1948distribution}\black{. Accordingly,} the pre-trained task \black{is related to} our target tasks. Moreover, as the input of our target tasks, we are given \black{the} ground-truth radar reflectivity for every pixel, while ground-truth precipitation can be observed only for a few pixels whose corresponding region\black{s have} weather stations.

\subsubsection{Fine-tuning phase}
\label{sec:method:training:tune}
In the fine-tuning phase, we train the pre-trained model (i.e., the model initialized with the parameters from the previous phase) for each of the target tasks described in Section~\ref{sec:problemdef}.
\blue{For precipitation nowcasting, we aim to minimize the difference  between (a) the output probability distribution over the three classes \heavyrain, \lightrain, and \norain for each pixel whose corresponding region has a weather station and (b) the ground-truth class distribution obtained from the precipitation measured at the weather stations.\footnote{Note that all probability masses of a ground-truth class distribution are assigned as the ground-truth class.}
While the cross-entropy (CE) loss (see Appendix~\ref{sec:app:loss}) is widely used for measuring such a difference, it is known to be affected by the class imbalance problem \citep{lin2017focal}. }
That is, deep-learning models aiming to minimize CE are easily biased toward the class \norain, which has many examples, and thus they have poor predictive performance for the classes \lightrain and \heavyrain, which have relatively fewer examples.

\black{To mitigate} the class imbalance problem, we design a new loss function inspired by the critical success index (CSI), which is defined for each class $c\in\{\lightrain, \heavyrain\}$ as 
\begin{equation}
    CSI_c = \frac{TP_{c}}{TP_{c} + FP_{c} + FN_{c}},  \label{eq:csi}
\end{equation}
where $TP_{c}$, $FP_{c}$, and $FN_{c}$ \black{represent} the numbers of true positives, false positives, and false negatives, respectively.
Note that being biased toward a major class tends to increase $FN_{c}$ in the denominator for a minor class $c$ and thus decrease the CSI score for $c$.
Thus, by aiming to increase the CSI score for minor classes, we can prevent deep-learning models from being biased toward major classes.

The CSI score cannot be directly used as a part of \black{the loss function as} it is not differentiable.
For differentiability, we use the probability distributions from a deep-learning model to approximate $TP_{c}$, $FP_{c}$, $FN_{c}$ for each class $c$, as in \cite{kaggle}.
We use $\widetilde{TP}_{c}$, $\widetilde{FP}_{c}$, and $\widetilde{FN}_{c}$ to denote the approximations of $TP_{c}$, $FP_{c}$, $FN_{c}$, respectively.
Suppose we approximate the CSI score for a class $c$, \black{and the predicted probability $\hat{P}_{t, i, c} = 0.8$.}
If the ground-truth class $P_{t, i}$ is $c$, then $\widetilde{TP}_c$ is increased by $0.8$, and $\widetilde{FN}_c$ is increased by $0.2$. 
If the ground-truth class $P_{t, i}$ is not $c$, $\widetilde{FP}_c$ is increased by $0.8$.
We increase $\widetilde{TP}_{c}$, $\widetilde{FP}_{c}$, and $\widetilde{FN}_{c}$ in the same manner for every predicted probability distribution, and as a result, obtain their final values.
Note that $\widetilde{TP}_{c}$, $\widetilde{FP}_{c}$, and $\widetilde{FN}_{c}$ are differentiable, and can \black{thus} be used \black{as} a part of \black{the loss function} for training deep-learning models.

As our final loss function for precipitation nowcasting, we use the negative average of the approximated CSI scores for minor classes as follows:
\begin{equation}
    L_{nowcasting} =  -\frac{1}{2}\left(\frac{\widetilde{TP}_{\lightheavyrain}}{\widetilde{TP}_{\lightheavyrain}+\widetilde{FP}_{\lightheavyrain}+\widetilde{FN}_{\lightheavyrain}}+\frac{\widetilde{TP}_{\heavyrain}}{\widetilde{TP}_{\heavyrain}+\widetilde{FP}_{\heavyrain}+\widetilde{FN}_{\heavyrain}}\right), \label{eq:loss:nowcasting}
\end{equation}
where $\lightheavyrain$ is a class defined as the union of $\lightrain$ and $\heavyrain$.
That is, an example belongs to  $\lightheavyrain$ if it belongs to $\lightrain$ or $\heavyrain$.
Due to the negative sign in the loss function above, aiming to minimize it \black{increases} the approximated CSI scores. Note that higher CSI scores indicate better classification performance\black{s}.
For mini-batch training, we \black{initially} compute $\widetilde{TP}_c$, $\widetilde{FP}_c$, and $\widetilde{FN}_c$ considering all cases in the considered batch and then calculate the loss based on them, instead of averaging the loss computed for each case.

For precipitation \black{estimation, we} use the widely-used sum squared error loss function (SSE) to measure the discrepancy between the ground-truth and \black{the} estimated precipitation, which are both real values, as follows:
\begin{equation}
    L_{estimation} = \sum_{R_{t}\in D} \sum_{i\in I'}\left(\hat{C}_{t,i} - C_{t,i}\right)^{2}, \label{eq:loss:estimation}
\end{equation}
where $D$ is  the set of radar images in the training set, and $I'\subseteq I$ is the set of pixels whose corresponding regions have weather stations.
\black{Because} the output dimensions of this fine-tuning task \black{differ from those} of the pre-training task, the parameters of the last convolutional layer are initialized randomly. The other parameters are initialized using their pre-trained values.

\section{Results}
\label{sec:experiments}
\noindent We review the experiments that we performed to answer the following questions:
\begin{enumerate}[start=1,label={\bfseries Q\arabic*.}]
    \item {\bf Effectiveness of our pre-training scheme:} How much does our pre-training scheme improve the accuracy of our deep-learning model? 
    \item {\bf Effectiveness of our loss function:} How much does our loss function (i.e., Eq.~\eqref{eq:loss:nowcasting}) improve the accuracy of our deep-learning model, compared to commonly-used loss functions? 
    \item {\bf Sensitivity to \black{the} resolution:} How does the accuracy of our deep-learning model depend on the resolution\black{s} of \black{the} inputs and outputs?
    \item {\bf Comparison with ZR relationships:}
    How accurate is our deep-learning model, compared to fitted ZR relationships  for precipitation estimation\black{s}?
\end{enumerate}

\subsection{Experimental settings}
\label{sec:experiments:settings}

We describe our experimental settings, including computational resources, data, and evaluation metrics.

\paragraph{Computational resources:} We performed all experiments on a server with 512GB \black{of} RAM and eight RTX 8000 GPUs, each of which has 48GB \black{of} GPU memory.

\begin{figure}
    \centering
    \includegraphics[trim={1.7cm .75cm 1.3cm .7cm},clip,width=0.45\linewidth]{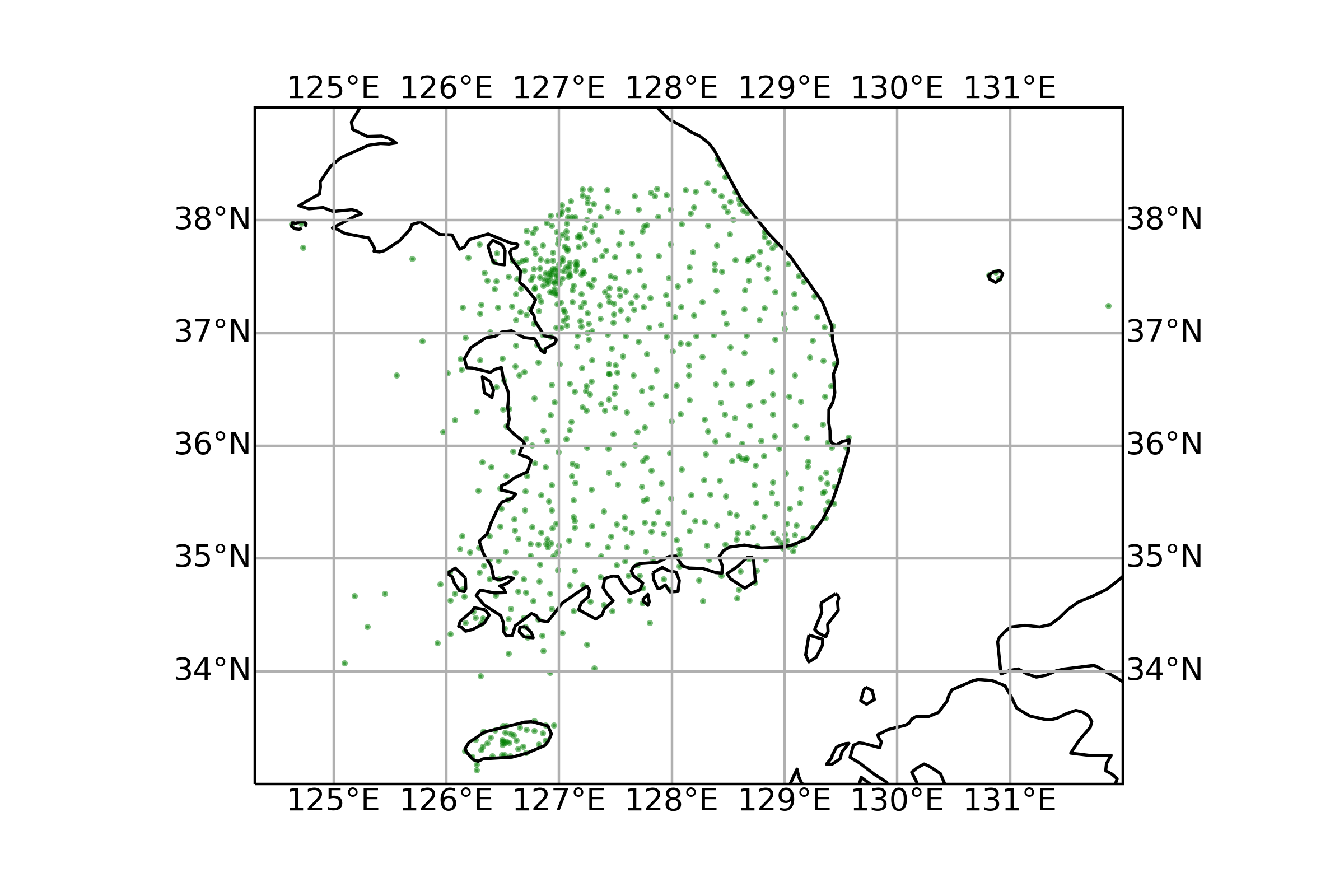}
    \includegraphics[trim={1.7cm .75cm 1.3cm .7cm},clip,width=0.45\linewidth]{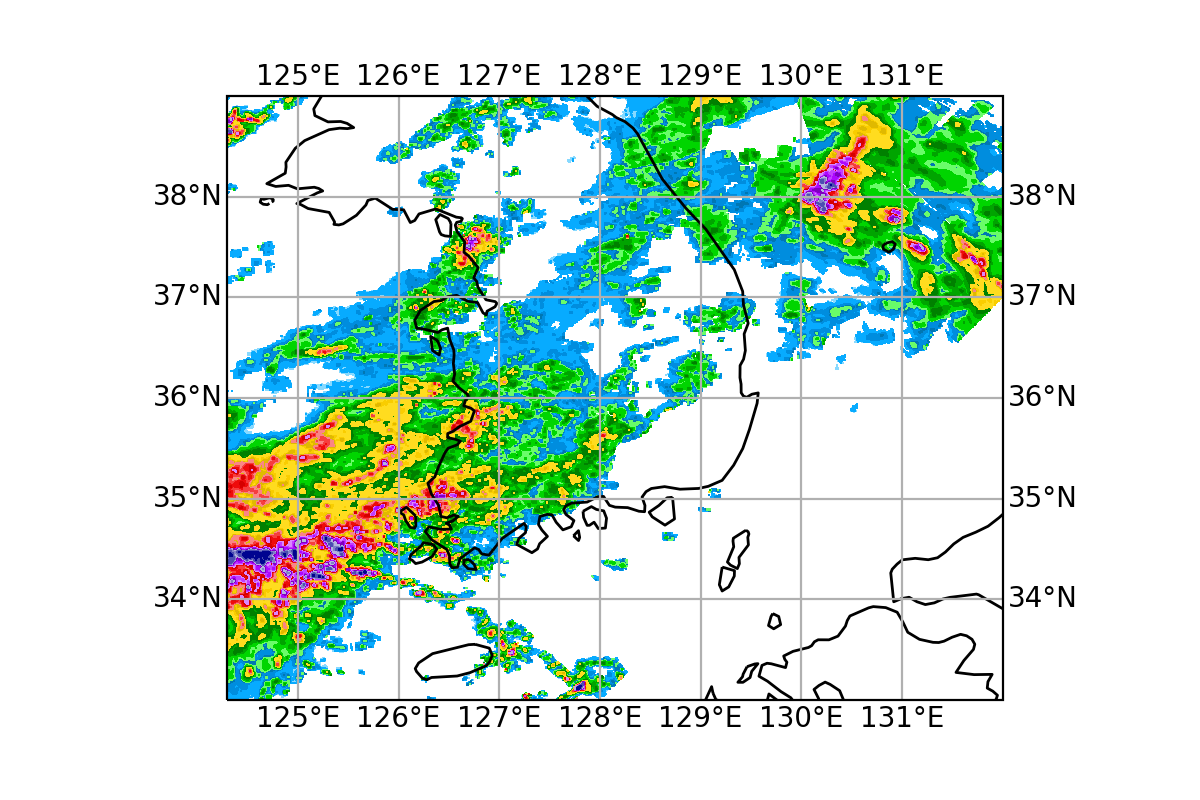}
    \caption{An example \black{of a} radar reflectivity image around South Korea, and the locations of weather stations over South Korea.}
    \label{fig:awslocations}
\end{figure}

\paragraph{Data:} We used radar reflectivity images around South Korea (specifically, longitude 120.5\textdegree E - 137.5\textdegree E and latitude 29.0\textdegree N - 42.0\textdegree N) that were measured every \black{ten} minutes from 2014 to 2020. We also used the amount of precipitation measured at $714$ automated weather stations (AWS) installed in South Korea.
The amount of precipitation was measured every minute \black{in each case}.
See Figure \ref{fig:awslocations} for the locations of the weather stations. 
\black{In the pre-training phase, we used all available radar images from 2014 to 2018 for training and those in 2019 validation}.
\black{In the fine-tuning phase,  we only used data in summer (from June to September), when most rain falls in South Korea,}
Specifically, we used radar images and accumulated precipitation from 2014 to 2018 for fine-tuning, those in 2019 for validation, and those in 2020 for evaluation.

\paragraph{Evaluation metrics:} For the evaluation of the task of precipitation nowcasting, we measured the CSI scores and F1 scores for two classes of rainfall events: \lightheavyrain (at least $1$ mm/hr) and \heavyrain (at least $10$ mm/hr) for the test set (2020 summer).
\footnote{\black{The CSI score is widely used \black{for} verifying weather forecasts \black{because} it is not affected by the number of non-event forecasts.
The F1 score is used commonly for deep\black{-}learning-based precipitation nowcasting.
We measured both for a comparison of recent studies on deep\black{-}learning-based precipitation nowcasting in Section~\ref{sec:discussion}.}}
The F1 score for class $c$ of rainfall events is defined as follows:
\begin{equation}
    F1_c = \frac{2\times TP_{c}}{2\times TP_{c} + FP_{c} + FN_{c}}.
\end{equation}
For the evaluation of the task of precipitation estimation, we measured the mean-squared error (MSE) between ground-truth accumulated precipitations and predicted ones as follows:
\begin{equation}
    MSE = \frac{1}{|D_{test}|}\sum_{R_{t}\in D_{test}}  \left(\frac{1}{|I'|} \sum_{i\in I'}\left(\hat{C}_{t,i} - C_{t,i}\right)^{2} \right),
\end{equation}
where $D_{test}$ is the set of radar images in the test set, and $I'\subseteq I$ is the set of pixels whose corresponding regions \black{have a weather station where ground-truth accumulated precipitation is measured.}
Note that higher F1 and CSI scores indicate better predictive performance\black{s} while a lower MSE indicates better predictive performance.

\paragraph{Default settings:}
Unless otherwise stated, we went through both pre-training and fine-tuning phases, as described in Section~\ref{sec:method:training}, and the inputs and outputs of our model were at a spatial resolution of $1$ km $\times$ $1$ km.
When the spatial resolution is $1$ km $\times$ $1$ km, the input and output dimensions are $1,468 \times 1,468$ and $706 \times 706$, respectively.

\paragraph{\blue{Details of training and validation}:}
\blue{We used the Adam optimizer \citep{kingma2014adam} with a learning rate of $2 \cdot 10^{-5}$ for both the pre-training and fine-tuning phases.
We trained our model for $50,000$ steps with a batch size of 20 in the pre-training phase and $35,000$ steps with a batch size of 24 in the fine-tuning phase.}
For every $1,000$ steps, we measured the overall performance (specifically, Eq.~\eqref{eq:loss:pretrain} in the pre-training phase and the CSI score for the class \heavyrain in the fine-tuning phase) in the validation dataset. 
At the end of training, we used the parameter values when the overall performance in the validation was maximized. Without pre-training, the overall performance is maximized after 5,000 steps, and in every setting, it is maximized within 28,000 steps.

\begin{table}[ht]
    \centering
    \caption{Effectiveness of our pre-training scheme in terms of predictive performance on the test set.
    In each setting, the best score is in bold. 
    \black{The persistence heuristic is the trivial identity model where the class of each location is predicted to be \black{identical to} the current class.}
    Our pre-training scheme significantly and consistently improved the predictive performance of our deep-learning model for precipitation nowcasting.
    }
    \begin{subtable}{0.49\textwidth}
        \centering
        \scalebox{0.9}{
        \hspace{-2mm}
        \renewcommand{\tabcolsep}{1.45mm}
        \begin{tabular}{c|cc|cc|cc}
            \toprule
            \multirow{2}{*}{\makecell{Lead time}} & \multicolumn{2}{c|}{\makecell{With \\ pre-training} } & \multicolumn{2}{c|}{\makecell{Fine-tuning\\ only}} & \multicolumn{2}{c}{\black{Persistence}}\\
            \cline{2-7}\rule{0pt}{2.5ex} & CSI & F1 & CSI & F1 & CSI & F1 \\ \hline \hline
            60 minutes 	&	\textbf{0.390} 	&	\textbf{0.562} 	&	0.376 	&	0.547 	&   0.259   &   0.412   \\
            120 minutes    &	\textbf{0.280} 	&	\textbf{0.438} 	&	0.225 	&	0.368 	&   0.152   &   0.264   \\
            180 minutes 	&	\textbf{0.210}	&	\textbf{0.348} 	&	0.179 	&	0.304 	&   0.102   &   0.185   \\
            240 minutes 	&	\textbf{0.170} 	&	\textbf{0.291} 	&	0.137 	&	0.240 	&   0.073   &   0.136   \\
            300 minutes 	&	\textbf{0.135} 	&	\textbf{0.238} 	&	0.069 	&	0.129 	&   0.057   &   0.108   \\
            360 minutes 	&	\textbf{0.116} 	&	\textbf{0.207} 	&	0.005 	&	0.009 	&   0.046   &   0.087   \\
            \midrule
            Average  	&	\textbf{0.217} 	&	\textbf{0.347} 	&	0.165 	&	0.266 	&   0.115   &   0.199   \\

            \bottomrule
        \end{tabular}
        }
        \caption{CSI/F1 score for prediction for \heavyrain ($\geq 10$ mm/hr)}
    \end{subtable}
    \begin{subtable}{0.49\textwidth}
        \centering
        \scalebox{0.9}{
        \hspace{-2mm}
        \renewcommand{\tabcolsep}{1.45mm}
        \begin{tabular}{c|cc|cc|cc}
            \toprule
            \multirow{2}{*}{\makecell{Lead time}} & \multicolumn{2}{c|}{\makecell{With \\ pre-training} } & \multicolumn{2}{c|}{\makecell{Fine-tuning\\ only}} & \multicolumn{2}{c}{\black{Persistence}} \\
            \cline{2-7}\rule{0pt}{2.5ex} & CSI & F1 & CSI & F1  & CSI & F1 \\ \hline \hline
            60 minutes 	&	\textbf{0.609} 	&	\textbf{0.757} 	&	0.600	&	0.750 	&   0.518   &   0.683   \\
            120 minutes 	&	\textbf{0.501} 	&	\textbf{0.667} 	&	0.497 	&	0.664 	&   0.396   &   0.568   \\
            180 minutes 	&	\textbf{0.449} 	&	\textbf{0.620} 	&	0.438 	&	0.609 	&   0.331   &   0.498   \\
            240 minutes 	&	\textbf{0.411} 	&	\textbf{0.583} 	&	0.407 	&	0.579 	&   0.288   &   0.447   \\
            300 minutes 	&	\textbf{0.381} 	&	\textbf{0.552} 	&	0.356 	&	0.525 	&   0.256   &   0.408   \\
            360 minutes 	&	\textbf{0.354} 	&	\textbf{0.523} 	&	0.261 	&	0.414 	&   0.231   &   0.375   \\
            \midrule
            Average  	&	\textbf{0.451} 	&	\textbf{0.617} 	&	0.426 	&	0.590 	&   0.337   &   0.496   \\

            \bottomrule
        \end{tabular}
        }
        \caption{CSI/F1 score for prediction for \lightheavyrain ($\geq 1$ mm/hr)}
    \end{subtable}
    \label{tab:pretrain}
\end{table}

\begin{table}[ht]
    \centering
    \caption{Effectiveness of our loss function (i.e., Eq.~\eqref{eq:loss:nowcasting}) in terms of predictive performance on the test set. We compared ours with the focal loss (Focal) \citep{lin2017focal} and the cross entropy loss (CE).
    As shown in (a), using ours led to the highest CSI and F1 scores for the class \heavyrain ($\geq 10$ mm/hr) when the lead time (i.e., $t^*-t$) was two hours or longer.
    }
    \begin{subtable}{0.49\textwidth}
        \scalebox{0.9}{
        \begin{tabular}{c|c@{\hspace{5pt}}c@{\hspace{5pt}}c|c@{\hspace{5pt}}c@{\hspace{5pt}}c}
            \toprule
            \multirow{2}{*}{Lead time} & \multicolumn{3}{c|}{CSI Score} & \multicolumn{3}{c}{F1 Score} \\
            \cline{2-7}\rule{0pt}{2.5ex} & Ours & Focal & CE & Ours & Focal & CE \\ \hline \hline
            60 minutes &  0.390 &	0.402 &	\textbf{0.422} &	0.562 &	0.574 &	\textbf{0.594} \\
            120 minutes & \textbf{0.280} &	0.255 &	0.257 &	\textbf{0.438} &	0.407 &	0.409 \\ 
            180 minutes & \textbf{0.210} &	0.187 &	0.150 &	\textbf{0.348} &	0.316 &	0.261 \\
            240 minutes & \textbf{0.170} &	0.130 &	0.095 &	\textbf{0.291} &	0.230 &	0.173 \\
            300 minutes & \textbf{0.135} &	0.094 &	0.061 &	\textbf{0.238} &	0.173 &	0.114 \\
            360 minutes & \textbf{0.116} &	0.081 & 0.031 &	\textbf{0.207} &	0.150 &	0.061 \\
            \midrule
            Average     & \textbf{0.217} &	0.192 & 0.169 &	\textbf{0.347} &	0.308 &	0.269 \\
            \bottomrule
        \end{tabular}
        }
        \caption{CSI/F1 score for \heavyrain ($\geq 10$ mm/hr)}
    \end{subtable}
    \label{tab:lossfn}
    \begin{subtable}{0.49\textwidth}
        \scalebox{0.9}{
        \begin{tabular}{c|c@{\hspace{5pt}}c@{\hspace{5pt}}c|c@{\hspace{5pt}}c@{\hspace{5pt}}c}
            \toprule
            \multirow{2}{*}{Lead time} & \multicolumn{3}{c|}{CSI Score} & \multicolumn{3}{c}{F1 Score} \\
            \cline{2-7}\rule{0pt}{2.5ex} & Ours & Focal & CE & Ours & Focal & CE \\ \hline \hline
            60 minutes  & 0.609             & 0.639             & \textbf{0.641}    & 0.757             & 0.780             & \textbf{0.782} \\
            120 minutes & 0.500             & \textbf{0.531}    & 0.528             & 0.667             & \textbf{0.694}    & 0.691 \\
            180 minutes & 0.449             & \textbf{0.456}    & 0.447             & 0.620             & \textbf{0.627}    & 0.618 \\
            240 minutes & \textbf{0.411}    & 0.404             & 0.398             & \textbf{0.583}    & 0.575             & 0.569 \\
            300 minutes & \textbf{0.381}    & 0.368             & 0.360             & \textbf{0.552}    & 0.538             & 0.529 \\
            360 minutes & \textbf{0.354}    & 0.346             & 0.327             & \textbf{0.523}    & 0.514             & 0.493 \\
            \midrule
            Average     & 0.451             & \textbf{0.457}    & 0.450             & 0.617             & \textbf{0.621}    & 0.614 \\
            \bottomrule
        \end{tabular}
        }
        \caption{CSI/F1 score for \lightheavyrain ($\geq 1$ mm/hr)}
    \end{subtable}
\end{table}

\subsection{Precipitation nowcasting}
\label{sec:experiments:nowcasting}

Below, we report and discuss the results of our experiments regarding the problem of precipitation nowcasting, \black{as} described in Section~\ref{sec:problemdef:nowcasting}.

\paragraph{Effectiveness of pre-training}
In order to demonstrate the effectiveness of our pre-training scheme, we compared the predictive performances of our deep-learning model with and without the pre-training phase described in Section~\ref{sec:method:training}.
\black{We additionally consider the persistence heuristic, which is the trivial identity model where the class of each location is predicted to be \black{identical to} the current class, as a baseline approach.}
We report the CSI and F1 scores for the classes \lightheavyrain ($\geq1$ mm/hr) and \heavyrain ($\geq10$ mm/hr) on the test set in Table~\ref{tab:pretrain}. \black{Additionally, we show the learning curves in Figure~\ref{fig:training_curve} and provide the confusion matrices in Appendix~\ref{sec:app:confusion}.}
We note that pre-training consistently improved the predictive performances of our deep-learning model.
For example, \black{as shown in Table~\ref{tab:pretrain},} pre-training improved the CSI and F1 scores for the class \heavyrain by $\mathbf{95.7\%}$ and $\mathbf{84.4\%}$, respectively, \black{compared to the fine-tuning only,} when the lead time (i.e., $t^*-t$) was five hours. 

\paragraph{Effectiveness of our loss function}
We compared the CSI \black{and} F1 scores of our deep-learning model with three different loss functions: our loss function (i.e., Eq.~\eqref{eq:loss:nowcasting}), the focal loss \citep{lin2017focal}, and the cross entropy loss. 
Note that the cross entropy loss is \black{widely used} for classification tasks, and the focal loss is known to be effective when the distribution of classes is skewed.
\black{The cross entropy loss and the focal loss are described in detail with equations in Appendix~\ref{sec:app:loss}.}
For the focal loss, we set $\gamma$ to $2$, which led to the best predictive performance, \black{among $1$, $2$, and $5$, on both the validation set and the test set.}
We report the CSI and F1 scores for the classes \lightheavyrain ($\geq 1$ mm/hr) and \heavyrain ($\geq 10$ mm/hr) on the test set in Table~\ref{tab:lossfn}.
As \black{shown} in Table~\ref{tab:lossfn},
for the class \lightheavyrain ($\geq 1$ mm/hr), there was no clear winner.
However, for the class \heavyrain ($\geq 10$ mm/hr), which is rarer than \lightheavyrain, using our loss function improved both \black{the} CSI and F1 scores on the test set consistently when the lead time (i.e., $t^*-t$) was two hours or longer.
For example, when the lead time was five hours, the improvements in terms of the CSI and F1 scores \black{were approximately} $\mathbf{43.6\%}$ and $\mathbf{38.0\%}$\black{, respectively, compared to the focal loss.}

\begin{figure}[H]
    \centering
    \vspace{-3mm}
    \includegraphics[width=0.7\linewidth]{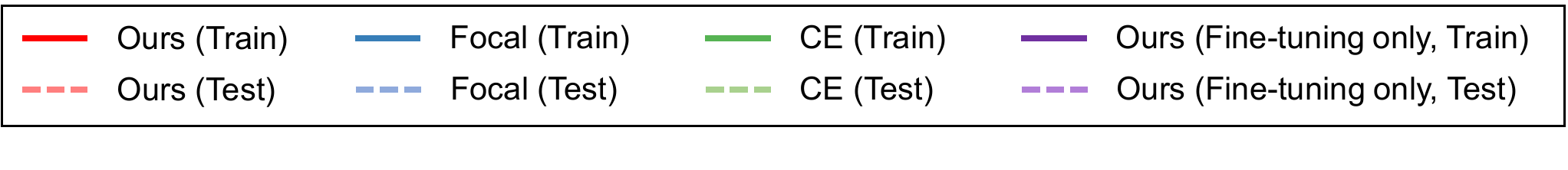} 
    \begin{subfigure}[]{0.43\linewidth}
        \includegraphics[width=\linewidth]{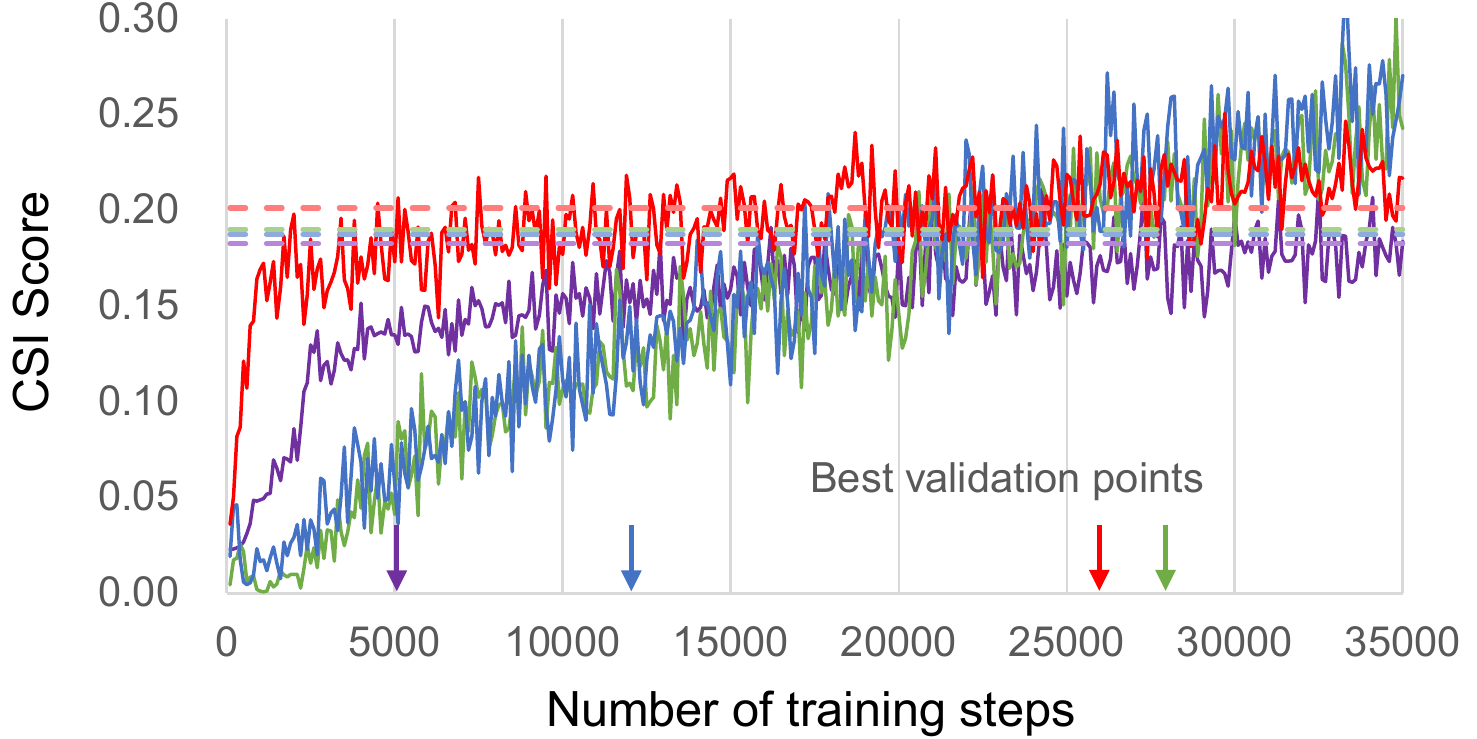}
        \caption{Training curve for \heavyrain ($\geq 10$ mm/hr)}
    \end{subfigure}
    \begin{subfigure}[]{0.43\linewidth}
        \includegraphics[width=\linewidth]{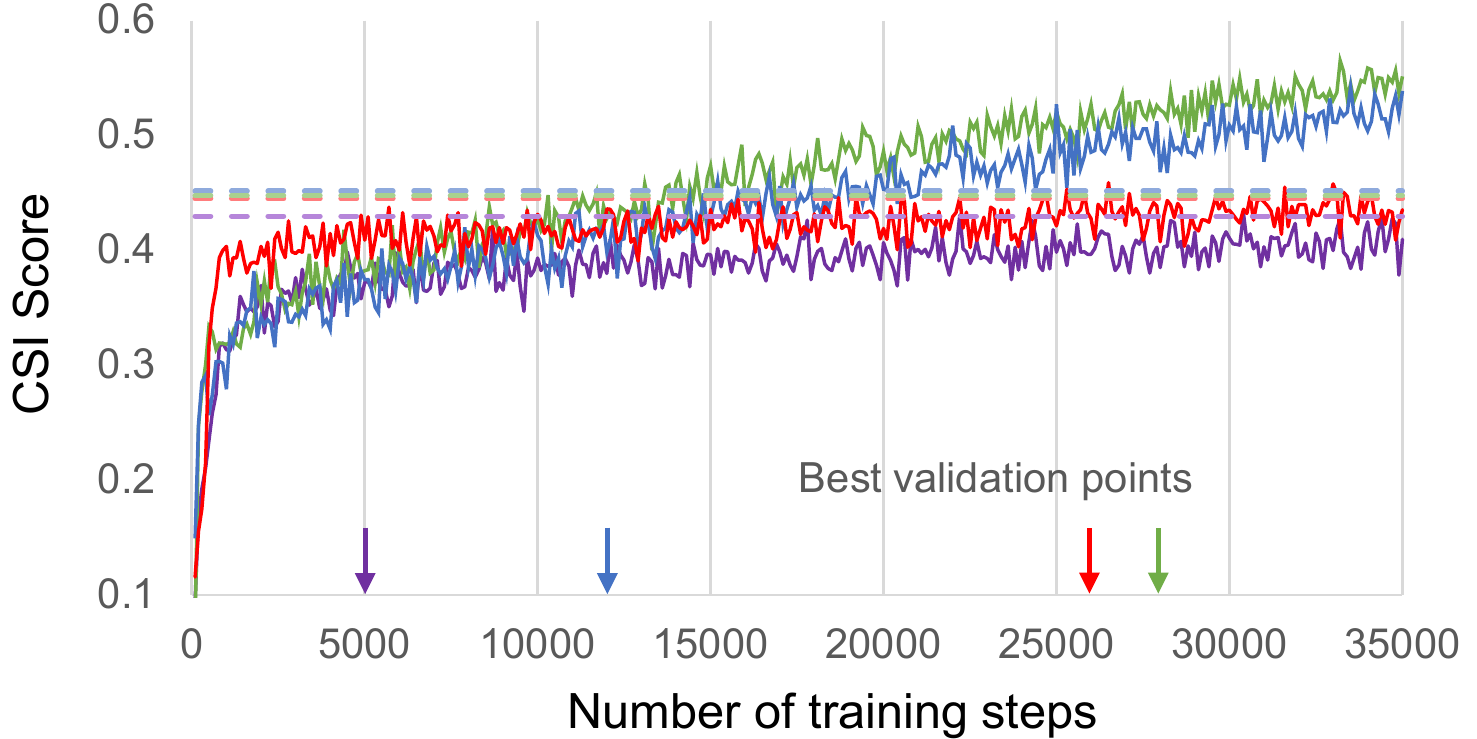}
        \caption{Training curve for \lightheavyrain ($\geq 1$ mm/hr)}
    \end{subfigure}
    \caption{\black{The training curves of our deep-learning model in four settings: (1) using our loss function (Eq.~\eqref{eq:loss:nowcasting}) with pre-training (\textcolor{red}{red}), (2) using our loss function without pre-training (\textcolor{violet}{purple}), (3) using the focal loss \citep{lin2017focal} (Focal) with pre-training (\textcolor{RoyalBlue}{blue}), and (4) using the cross entropy loss (CE) with pre-training (\textcolor{ForestGreen}{green}).
    Solid lines represent the overall CSI scores in the training set at each training step, and dashed lines represent the overall CSI scores in the test set at the best validation points, which are indicated by arrows \black{here}.    
    The overall CSI scores are computed from the confusion matrix for all training or test cases at all lead times.
    Note that when both our pre-training scheme and our loss function are used, the CSI score increase\black{s} rapidly, especially for the class \heavyrain ($\geq 10$ mm/hr).
   \black{The h}igh CSI scores in the training set achieved after about $20,000$ steps were due to overfitting, and they thus did not imply high CSI scores in the test set, which are reported in Table~\ref{tab:lossfn}.}
    }
    \label{fig:training_curve}
\end{figure}
\begin{table}[H]
    \centering
    \caption{Sensitivity to resolution. In (a) and (b), we reported the CSI/F1 scores of our deep-learning model at resolutions $1$ km $\times$ $1$ km, $2$ km $\times$ $2$ km, and $4$ km $\times$ $4$ km, on the test set.
    Note that the highest resolution (i.e., $1$ km $\times$ $1$ km) was generally helpful for the class \heavyrain ($\geq 10$ mm/hr).
    In (c), we reported the number of trainable parameters of our model at each resolution.
    }
    \begin{subtable}{0.49\textwidth}
        \centering
        \scalebox{0.9}{
        \begin{tabular}{c|c@{\hspace{5pt}}c@{\hspace{5pt}}c|c@{\hspace{5pt}}c@{\hspace{5pt}}c}
            \toprule
            \multirow{2}{*}{Lead time} & \multicolumn{3}{c|}{CSI Score} & \multicolumn{3}{c}{F1 Score} \\
            \cline{2-7}\rule{0pt}{2.5ex} & 1$\times$1 & 2$\times$2 & 4$\times$4 & 1$\times$1 & 2$\times$2 & 4$\times$4 \\ \hline \hline
            60 minutes	&	0.390 	&	\textbf{0.406} 	&	{0.400} 	&	0.562 	&	\textbf{0.578} 	&	{0.571} 	\\
            120 minutes	&	\textbf{0.280} 	&	0.254 	&	0.261 	&	\textbf{0.438} 	&	0.405 	&	0.413 	\\
            180 minutes	&	\textbf{0.210} 	&	0.177 	&	0.196 	&	\textbf{0.348} 	&	0.301 	&	0.328 	\\
            240 minutes	&	\textbf{0.170} 	&	0.130 	&	0.146 	&	\textbf{0.291} 	&	0.231 	&	0.254 	\\
            300 minutes	&	\textbf{0.135} 	&	0.104 	&	0.102 	&	\textbf{0.238} 	&	0.188 	&	0.186 	\\
            360 minutes	&	\textbf{0.116} 	&	0.077 	&	0.076 	&	\textbf{0.207} 	&	0.144 	&	0.142 	\\
            \midrule
            Average     & \textbf{0.217} &	0.191 & 0.197 &	\textbf{0.347} &	0.308 &	0.316 \\
            \bottomrule
        \end{tabular}
        }
        \caption{CSI/F1 score for \heavyrain ($\geq 10$ mm/hr)}
    \end{subtable}
    \begin{subtable}{0.49\textwidth}
        \centering
        \scalebox{0.9}{
        \begin{tabular}{c|c@{\hspace{5pt}}c@{\hspace{5pt}}c|c@{\hspace{5pt}}c@{\hspace{5pt}}c}
            \toprule
            \multirow{2}{*}{Lead time} & \multicolumn{3}{c|}{CSI Score} & \multicolumn{3}{c}{F1 Score} \\
            \cline{2-7}\rule{0pt}{2.5ex} & 1$\times$1 & 2$\times$2 & 4$\times$4 & 1$\times$1 & 2$\times$2 & 4$\times$4 \\ \hline \hline
            60 minutes	&	0.609 	&	0.594 	&	\textbf{0.620} 	&	{0.757} 	&	0.745 	&	\textbf{0.765} 	\\
            120 minutes	&	0.501 	&	\textbf{0.517} 	&	0.506 	&	{0.667} 	&	\textbf{0.682} 	&	0.672 	\\
            180 minutes	&	{0.449} 	&	\textbf{0.451} 	&	{0.448} 	&	{0.620} 	&	\textbf{0.622} 	&	{0.619} 	\\
            240 minutes	&	{0.411} 	&	{0.408} 	&	\textbf{0.415} 	&	{0.583} 	&	{0.579} 	&	\textbf{0.587} 	\\
            300 minutes	&	{0.381} 	&	{0.383} 	&	\textbf{0.388} 	&	{0.552} 	&	{0.553} 	&	\textbf{0.559} 	\\
            360 minutes	&	{0.354} 	&	\textbf{0.361} 	&	{0.360} 	&	{0.523} 	&	\textbf{0.531} 	&	{0.530} 	\\
            \midrule
            Average     & 0.451 &	0.452 & \textbf{0.456} &	0.617 &	0.619 &	\textbf{0.622} \\
            \bottomrule
        \end{tabular}
        }
        \caption{CSI/F1 score for \lightheavyrain ($\geq 1$ mm/hr)}
    \end{subtable}
    \begin{subtable}{0.99\textwidth}
        \centering
        \scalebox{0.9}{
        \begin{tabular}{c|c@{\hspace{5pt}}c@{\hspace{5pt}}c}
            \toprule
            Resolution &  1$\times$1 & 2$\times$2 & 4$\times$4 \\
            \hline \hline \rule{0pt}{2.5ex}
            Number of trainable parameters & 124.5M & 31.1M & 7.8M \\
            \bottomrule
        \end{tabular}
        }
        \caption{\blue{Number of trainable parameters of our deep-learning model}}
    \end{subtable}
    \label{tab:resolution}
\end{table}

\blue{We also provide the learning curves and the confusion matrices in Figure~\ref{fig:training_curve} and Appendix~\ref{sec:app:confusion}, respectively.
As shown in the confusion matrices, when our loss function was used with pre-training, there was a slight tendency of overestimation of precipitation. Specifically, it led to an overestimation in 5.41\% of the considered cases and an underestimation in 2.86\% of the considered cases.}

\begin{table}[ht]
    \caption{
    Performance analysis for \black{four cases} of heavy rainfall.
    For each \black{case}, we measured the CSI score of our deep-learning model at the stations near each site of heavy rainfall.
    We increased the maximum distance from any station where each heavy rainfall was observed from $10$ km to $50$ km.
    Note that the scores were relatively \black{low for Case 2, which is an isolated heavy rainfall case.}
    }
    \centering
    \scalebox{0.9}{
    \begin{tabular}{c@{\hspace{3pt}}|@{\hspace{3pt}}c@{\hspace{3pt}}|@{\hspace{3pt}}c@{\hspace{5pt}}c@{\hspace{5pt}}c@{\hspace{5pt}}c@{\hspace{5pt}}c@{\hspace{5pt}}c|c@{\hspace{5pt}}c@{\hspace{5pt}}c@{\hspace{5pt}}c@{\hspace{5pt}}c@{\hspace{5pt}}c@{\hspace{3pt}}}
        \toprule
         \multirow{2}{*}{Distance} & Class & \multicolumn{6}{c|}{ \black{\lightheavyrain (at least $1$ mm/hr)}} & \multicolumn{6}{c}{ \black{\heavyrain (at least $10$ mm/hr)}} \\
         \cline{2-14}\rule{0pt}{2.5ex} & Lead time (in minutes) & 60 & 120 & 180 & 240 & 300 & 360 & 60 & 120 & 180 & 240 & 300 & 360 
         \\ \hline\hline
         \multirow{4}{*}{10 km}	&	Case 1	&	1.000 	&	1.000 	&	1.000 	&	0.600 	&	1.000 	&	1.000 	&	1.000 	&	0.000 	&	0.400 	&	0.000 	&	0.000	&	1.000	\\ 
         &	Case 2	&	1.000 	&	1.000 	&	1.000 	&	0.000 	&	0.000 	&	0.000 	&	1.000 	&	1.000	&	1.000	&	0.000	&	0.000	&	0.000	\\
         &	Case 3	&	1.000 	&	1.000 	&	1.000 	&	1.000 	&	1.000 	&	0.000 	&	0.000 	&	0.000 	&	1.000 	&	1.000 	&	1.000	&	0.000 	\\ 
         &	Case 4	&	1.000 	&	1.000 	&	1.000 	&	1.000 	&	1.000 	&	0.000 	&	0.667 	&	1.000 	&	0.000 	&	0.334 	&	0.000	&	0.000	\\ 
         \midrule\multirow{4}{*}{25 km} &	Case 1	&	1.000 	&	1.000 	&	1.000 	&	0.618 	&	1.000 	&	1.000 	&	0.912 	&	0.061 	&	0.559 	&	0.000 	&	0.265	&	0.941	\\ 
         &	Case 2	&	1.000 	&	1.000 	&	1.000 	&	0.000 	&	0.000 	&	0.000 	&	1.000 	&	0.714	&	0.714	&	0.000	&	0.000	&	0.000	\\ 
         &	Case 3	&	1.000 	&	1.000 	&	1.000 	&	0.917 	&	1.000 	&	0.334 	&	0.583 	&	0.000 	&	1.000 	&	0.583 	&	1.000	&	0.000	\\ 
         &	Case 4	&	1.000 	&	1.000 	&	1.000 	&	1.000 	&	1.000 	&	0.077 	&	0.692 	&	1.000 	&	0.154 	&	0.462 	&	0.000	&	0.000	\\ 
         
	\midrule\multirow{4}{*}{50 km}	&	Case 1	&	1.000 	&	1.000 	&	1.000 	&	0.655 	&	0.876 	&	1.000 	&	0.776 	&	0.320 	&	0.518	&	0.055	&	0.306	&	0.726	\\ 
	&	Case 2	&	0.758 	&	0.758 	&	0.758 	&	0.267 	&	0.000	&	0.000	&	0.750 	&	0.476	&	0.438	&	0.000	&	0.000	&	0.000	\\ 
	&	Case 3	&	1.000 	&	1.000 	&	1.000 	&	0.625 	&	1.000 	&	0.100 	&	0.625 	&	0.425 	&	0.625 	&	0.350 	&	0.949	&	0.000	\\ 
	&	Case 4	&	1.000 	&	1.000 	&	1.000 	&	1.000 	&	1.000 	&	0.161 	&	0.677 	&	1.000 	&	0.581	&	0.774 	&	0.000	&	0.000	\\ 
	
	\midrule &	Case 1	&	0.662 	&	0.639 	&	0.587 	&	0.366 	&	0.506 	&	0.637 	&	0.538 	&	0.328	&	0.436	&	0.089	&	0.263	&	0.567	\\ 
	All &	Case 2	&	0.455 	&	0.298 	&	0.273 	&	0.157	&	0.000	&	0.093	&	0.510	&	0.361	&	0.304	&	0.000	&	0.000	&	0.000	\\ 
	Stations &	Case 3	&	0.690 	&	0.585 	&	0.689 	&	0.498 	&	0.598 	&	0.028 	&	0.408 	&	0.380 	&	0.556	&	0.462	&	0.408	&	0.000	\\ 
	&	Case 4	&	0.413 	&	0.366 	&	0.367 	&	0.416 	&	0.341 	&	0.327	&	0.373 	&	0.667	&	0.261	&	0.316	&	0.000	&	0.000	\\
 \bottomrule

    \end{tabular}
    }
    \label{tab:rainfall}
\end{table}

\paragraph{Sensitivity to \black{the} resolution.}
We analyzed how the resolutions of the input and output of our deep-learning model affects its predictive performance on the test set.
\black{Note that the resolution of the input and that of the output are always the same, and
when decreasing the resolution of the input, we used $2$ $\times$ $2$ or $4$ $\times$ $4$ mean pooling}.
We measured the CSI and F1 scores of our deep-learning model at resolutions \black{of} $1$ km $\times$ $1$ km, $2$ km $\times$ $2$ km, and $4$ km $\times$ $4$ km; as reported in Table~\ref{tab:resolution}(c), we changed the dimensions (i.e., the number of trainable parameters) of our model accordingly.
As \black{presented} in Table~\ref{tab:resolution}, the highest resolution (i.e., $1$ km $\times$ $1$ km) was generally helpful for the class \heavyrain ($\geq 10$ mm/hr), while the effect of \black{the} resolution was marginal and inconsistent for the class \lightheavyrain ($\geq 1$ mm/hr).

\begin{figure}
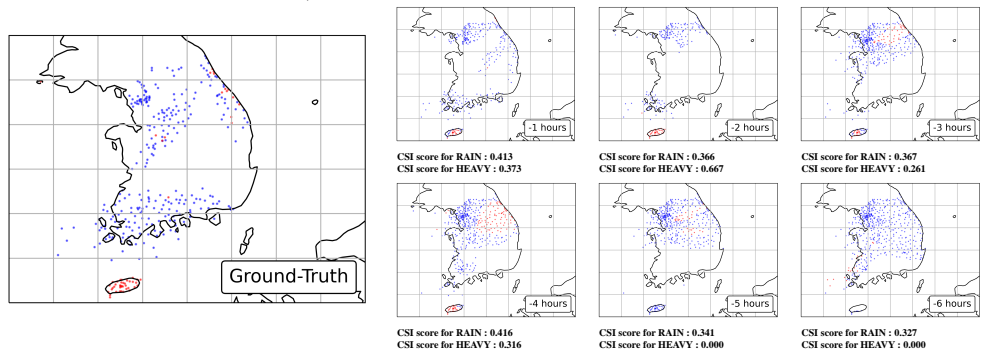

    \centering
    \textbf{Case 1} (August 6, 2020, 6:00 AM) \hfill ~ \\
    \includegraphics[width=0.8\linewidth]{figs/fig5_1.pdf} \\
    \textbf{Case 2} (August 2, 2020, 4:00 AM) \hfill ~ \\
    \includegraphics[width=0.8\linewidth]{figs/fig5_2.pdf} \\
    \textbf{Case 3} (August 7, 2020, 2:00 PM) \hfill ~ \\
    \includegraphics[width=0.8\linewidth]{figs/fig5_3.pdf} \\
    \textbf{Case 4} (September 2, 2020, 5:00 PM) \hfill ~ \\
    \includegraphics[width=0.8\linewidth]{figs/fig5_4.pdf}
    
    \caption{Prediction of \black{the} deep-learning model depending on \black{the} time when the prediction was made for examples of heavy rainfall cases. 
    We colored each weather station based on the predicted classes: {\color{red} red} for \heavyrain ($\geq 10$ mm/hr), {\color{blue} blue} for \lightrain ($1$ - $10$ mm/hr). \black{For each case, the large figure on the left shows the ground-truth precipitation, and the CSI scores are provided below the other figures.}
    }
    \label{fig:rainseqs_ver2}
\end{figure}

\paragraph{Performance analysis for four heavy rainfall cases.}
\black{We analyzed the predictive performance of our deep learning model for \black{four} spatially different heavy rainfall cases \black{with} precipitation intensity \black{levels exceeding} 30 mm/hr over South Korea on the test set. 
As shown in Figure~\ref{fig:rainseqs_ver2} (see the left penal in each row), Case 1 represents heavy rainfall observed widely in the central region of South Korea, and Case 2 represents heavy rainfall isolated in a small region. Case 3 represents heavy rainfall case\black{s} observed widely in the southern region of South Korea. Case 4 is similar to Case 3, but the maximum rainfall intensity was observed \black{on} Jeju Island. The \black{six} figures on the right side of each row in Figure~\ref{fig:rainseqs_ver2} show the spatial distribution of deep\black{-}learning-based heavy rainfall forecast according to the lead times. In general, deep-learning model-based forecast rainfalls for the lead time of 1 hour show very similar spatial distribution\black{s} to observations regardless of heavy rainfall cases. This indicates that our deep-learning model has a possibility to accurately predict heavy rainfall-affected regions 1 hour before an actual heavy rainfall event. 
However, the deep-learning model tends to over-predict heavy-rainfall-affected regions with an increase in the lead time for all heavy rainfall cases. For each heavy rainfall case, the CSI scores from the prediction results (at every weather station) are summarized in Table ~\ref{tab:rainfall}. In addition, in Table ~\ref{tab:rainfall}, the sensitivity results \black{regarding} the use of weather stations within various distance radii, i.e., 10, 25, 50 km, away from heavy rainfall events are also presented. Overall, the CSI scores are relatively low for Case 2. As mentioned above briefly, Case 2 is an isolated heavy rainfall case likely related to mesoscale convective activities. Such convective rainfalls intensely affect small regions within a short lifetime, \black{possibly} making \black{precise predictions difficult}.}

\subsection{Precipitation estimation}
\label{sec:experiments:estimation}

\begin{table}
    \centering
    \caption{
    Analysis for the task of precipitation estimation. Our deep-learning model tended to be more accurate with our pre-training scheme and at a higher resolution.
    For example, pre-training reduced the MSE on the test set by $\mathbf{6.06\%}$ at a $4$ km $\times$ $4$ km resolution.
    }
    \scalebox{0.9}{
    \begin{tabular}{c|c|c}
        \toprule
        Resolution & Training Method & MSE (the lower, the better) \\
        \hline\hline\rule{0pt}{2.5ex}
        \multirow{2}{*}{$1$ km $\times$ $1$ km} 	& With pre-training	& \textbf{1.362} \\ 
        & Fine-tuning only &	1.385  \\
        \midrule
        \multirow{2}{*}{$2$ km $\times$ $2$ km} 	& With pre-training	& \textbf{1.474} \\ 
        & Fine-tuning only &	1.518  \\
        \midrule
        \multirow{2}{*}{$4$ km $\times$ $4$ km} 	& With pre-training	& \textbf{1.457} \\ 
        & Fine-tuning only &	1.551  \\
        \bottomrule
    \end{tabular}
    }
    \label{tab:resolution_est}
\end{table}

Below, we review our experiments \black{on} the task of precipitation estimation, which is described in Section~\ref{sec:problemdef:estimation}. 

\paragraph{Sensitive to \black{the} resolution.}
We examined the sensitivity of our deep-learning model to \black{different} resolution\black{s}. To this end, we measured the MSE on the test set at three different resolutions ($1$ km $\times$ $1$ km, $2$ km $\times$ $2$, and $4$ km $\times$ $4$) after adjusting the number of trainable parameters accordingly. See Table~\ref{tab:resolution}(c) for the number of trainable parameters at each resolution. As \black{presented} in Table~\ref{tab:resolution_est}, our deep-learning model tended to be more accurate at a higher resolution.

\paragraph{Effectiveness of pre-training.}
We investigated how our pre-training scheme affected the estimation error of our deep-learning model for precipitation estimation.
As \black{shown} in Table~\ref{tab:resolution_est}, pre-training reduced the MSE on the test set consistently at every resolution.
For example, when the resolution was $4$ km $\times$ $4$ km, pre-training reduced the MSE by $\mathbf{6.06\%}$, \black{compared to the fine-tuning only}.

\begin{table}
    \centering
    \caption{Comparison with ZR relationships.
    Our deep-learning model for precipitation estimation was more accurate than a fitted ZR relationship for the class \lightrain ($1$ - $10$ mm/hr),
    while it was less accurate for the class \heavyrain ($\geq 10$ mm/hr), on the test set.
    }
    \scalebox{0.9}{
    \begin{tabular}{c|ccccc}
        \toprule
        \multirow{3}{*}{Method} & \multicolumn{3}{c}{MSE (the lower, the better)} \\
        \cline{2-4}\rule{0pt}{2.5ex} & \multirow{2}{*}{All} & \heavyrain  & \lightrain \\
        & & ($\geq 10$ mm/hr) & ($1$ - $10$ mm/hr) \\ 
        \hline\hline\rule{0pt}{2.5ex}
        ZR relationship	&	1.392 	&	\textbf{82.119} 	&	4.144  \\
         Ours	&	\textbf{1.362} 	&	87.215 	&	\textbf{3.699} 	\\
        \bottomrule
    \end{tabular}
    }
    \label{tab:zr}
\end{table}

\begin{figure}
    \centering
    \begin{subfigure}[]{0.3\linewidth}
        \centering
        \includegraphics[trim={2.5cm 1.1cm 1.3cm 1.2cm}, clip,width=\linewidth]{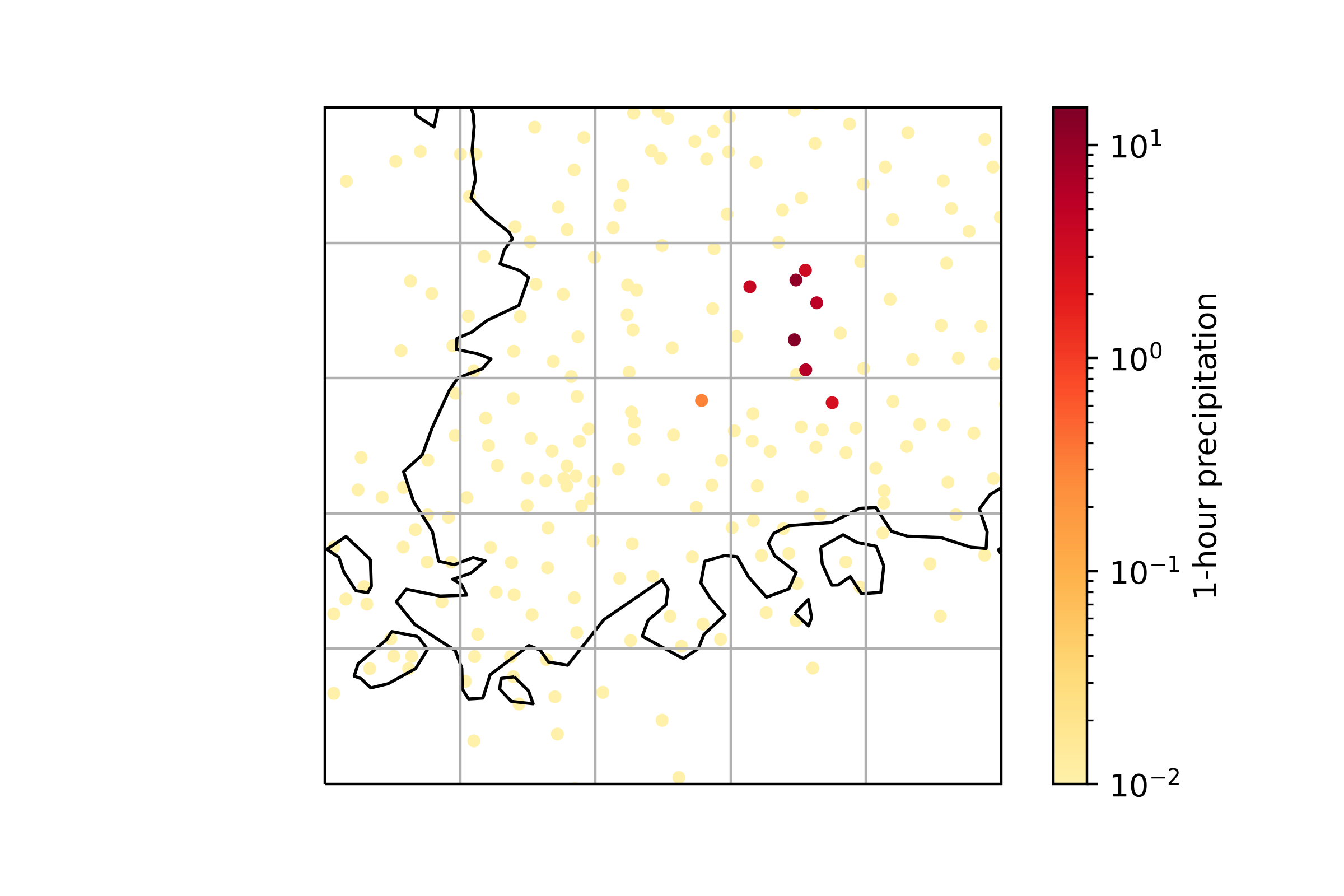}
        \caption{\small Ground-truth precipitation}
    \end{subfigure}
    \hspace{-2mm}
    \begin{subfigure}[]{0.3333\linewidth}
        \centering
        \includegraphics[trim={2.5cm 1.1cm 1.3cm 1.2cm}, clip,width=0.9\linewidth]{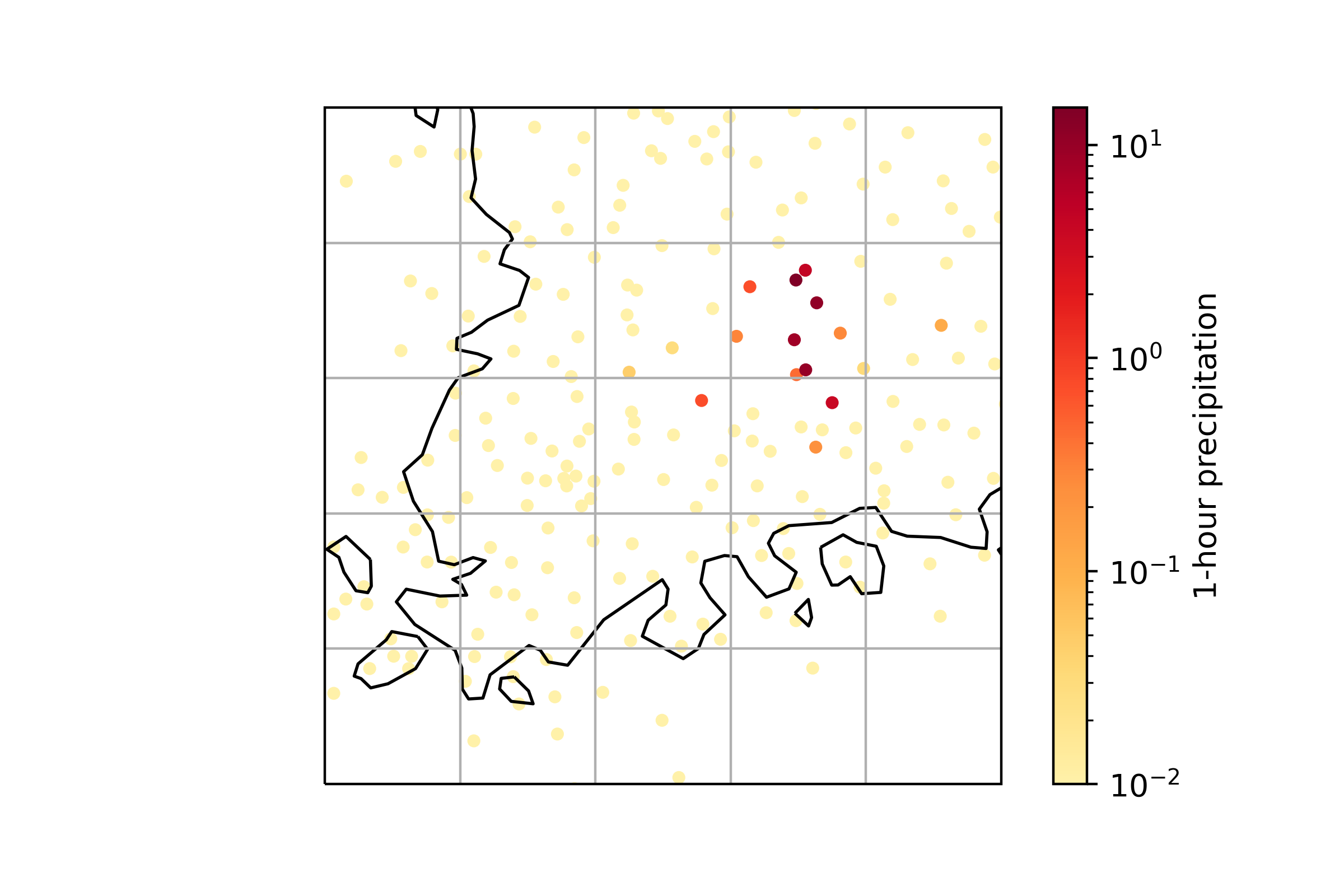}
        \caption{\small Estimation (deep-learning model)}
    \end{subfigure}
    \hspace{-2mm}
    \begin{subfigure}[]{0.3333\linewidth}
        \centering
        \includegraphics[trim={2.5cm 1.1cm 1.3cm 1.2cm}, clip,width=0.9\linewidth]{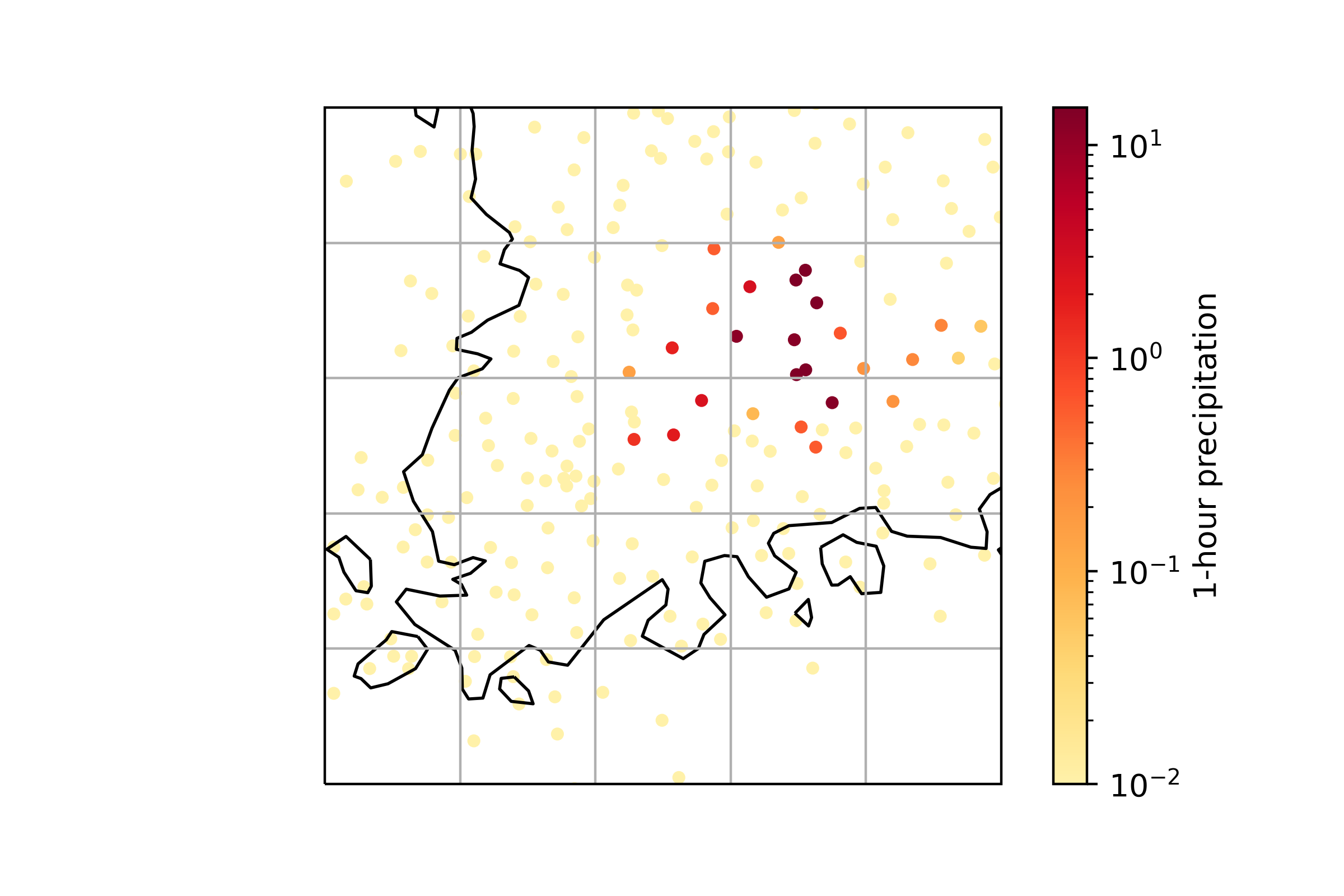}
        \caption{\small  Estimation (fitted ZR relationship)}
    \end{subfigure}
    \caption{(a) Ground-truth precipitation \black{\black{on} June 6, 2020, \black{at} 16:00 PM, (b) precipitation estimation by the deep-learning model, and (c) precipitation estimation by the fitted ZR relationship.}}
    \label{fig:compare_zr}
\end{figure}

\paragraph{Comparison with ZR relationships.} 
We compared our deep-learning model with ZR-relationship-based precipitation estimation.
A ZR relationship is a power-law relationship between radar reflectivity factor $Z$ and precipitation rate $R$. 
The relationship \black{takes} the form \black{of} $Z = aR^b$, and we set $a$ and $b$ \black{equal} $200$ and $1.49$, respectively, after fitting them to $Z$ and $R$ values around South Korea.
The fitted ZR relationship can naturally be used to estimate the precipitation rate at each region from a radar image\black{.} 
\black{We then} estimated the amount of one-hour precipitation by averaging the precipitation rates from the input radar images.

We measured the MSE of our deep-learning model and the fitted ZR relationship using all test cases and using those of each class. 
As \black{indicated} in Table~\ref{tab:zr}, our deep-learning model is comparable to the conventional ZR relationship\black{, and it achieves a }slightly smaller MSE for the classes \lightrain (between $1$ and $10$ mm/hr).
In Figure~\ref{fig:compare_zr}, for an example case, we visualize the estimation of our deep-learning model and the fitted ZR relationship with the ground-truth precipitation.

\section{Discussion}
\label{sec:discussion}

\black{As presented in Section~\ref{sec:experiments}, our training strategies significantly improved the predictive performance of deep learning-based precipitation nowcasting from radar images. Specifically, pre-training our deep-learning model for predicting radar reflectivity, which is generally correlated with precipitation rates, improved the average accuracy; and using our loss function improves the predictive performance for heavy rainfall (i.e., $\geq 10$ mm/hr) by mitigating the class imbalance problem. 
We compare the improvements in F1 and CSI scores reported in the recent deep-learning\black{-}based nowcasting in Table~\ref{tab:improve}. 
\black{Given that} the detailed settings (including areas, data sources, target precipitation rates, lead time, and target measures) vary, direct comparisons cannot be made. Nevertheless, the improvement  achieved by our approach for heavy rainfall is outstanding, especially when considering the fact that it relies only on radar images. Specifically, compared to the persistence heuristic, our approach improved the CSI score by \textbf{89\%} on average (by \textbf{137\%} at \black{a} 5-hr lead time) and the F1 score by \textbf{74\%} on average (by \textbf{120\%} at \black{a} 5-hr lead time), for heavy rainfall.
}

\black{A great advantage of our training schemes is that it is not restricted or specialized to a specific deep-learning model but can be applied to a wide range of deep-learning models. In this paper, however, we focus on an extended U-Net-based model (see Section~\ref{sec:method}) when verifying our training strategies. 
This means that our software inherits the limitation of the U-Net-based model. However, these limitations are partially addressed by recent studies described below.}

\black{While our deep-learning model solely relies on radar images, \cite{sonderby2020metnet} used satellite images, topographical data, and spatiotemporal embeddings in addition to radar images to improve the predictive performance. \cite{lebedev2019precipitation} used satellite images, GFS fields, solar altitude, and topographical data to reconstruct incomplete radar images with U-Net and then used the reconstructed images for precipitation nowcasting. 
Compared to the persistent heuristic, these approaches improved the F1 score by up to $86\%$ on average (by up to $112\%$ at \black{a} 5-hr lead time), 
as summarized in Table~\ref{tab:improve}.
Our deep-learning model can also be extended to incorporate such additional inputs, and specifically they can be added to the input tensor as additional channels.
}
\begin{table}[H]
    \centering
    \caption{\black{Comparison of recent studies \black{of} deep learning-based precipitation nowcasting conducted worldwide.
    For each study, we summarize the results in settings most similar to ours. 
    Note that the improvement for heavy rainfall \black{nowcasting} achieved by our approach, which uses only radar images, is outstanding.
    }}
        \renewcommand{\arraystretch}{2.5}
        \scalebox{0.87}{
        \begin{tabular}{c|c|c|c|c|c|c|c}
        \toprule
            Model   & Areas & \makecell{Data \\ Sources}  & \makecell{Precipitation\\ Rates}	& \makecell{Lead \\ Time}	& Measure	& Baseline	& \makecell{Improvement \\ over the Baseline} \\
			\hline \hline 
            \makecell{\unet\\\citep{agrawal2019machine}}  & US	& Radar   & $\geq 1mm $	& $1$ hr	& F1 score	&  Persistence	& $\approx 23\% $ \\\hline
            \makecell{MetNet\\\citep{sonderby2020metnet}}	& US	& \makecell{Radar, \\satellite, \\ and more}	& $\geq 1mm $	&       $1$ - $6$ hrs	& F1 score	&  Persistence	& \makecell{$\approx 86\%$ on average \\ ($\approx 112\%$ at 5-hr lead time)}
             \\\hline
            \makecell{\unet + CNN\\\citep{lebedev2019precipitation}}   & Russia	& \makecell{Radar, \\satellite, \\ and more}	& $\geq 0.08mm $	& $1$ hr	& F1 score	&  Persistence	& $\approx 22\% $ \\\hline
			\multirow{2}{*}{\makecell{DGMR\\\citep{ravuri2021skillful}}}   & UK & \multirow{2}{*}{Radar} & \multirow{2}{*}{$\geq 1mm $}   & \multirow{2}{*}{$1$ hr}  & \multirow{2}{*}{CSI score} & \multirow{2}{*}{\unet} & $\approx 0\% $ \\ 
			\cline{2-2}\cline{8-8}
               & US &   &    &  &  &  &  $\approx 0\% $ \\ \hline
               
			\multirow{8}{*}{Ours}   & \multirow{8}{*}{\makecell{South\\ Korea}}  & \multirow{8}{*}{Radar}    & \multirow{4}{*}{$\geq 1mm $} & \multirow{8}{*}{$1$ - $6$ hrs}    & \multirow{2}{*}{F1 score}   &  Persistence      & \makecell{$\approx 24\%$ on average \\ ($\approx 35\%$ at 5-hr lead time)}\\
			\cline{7-8}
			 &      &    &    &   &   & \unet*   & \makecell{$\approx 5\%$ on average \\ ($\approx 5\%$ at 5-hr lead time)}\\
			 \cline{6-8}
			 &      &    &    &   & \multirow{2}{*}{CSI score}     &  Persistence      & \makecell{$\approx 34\%$ on average \\ ($\approx 49\%$ at 5-hr lead time)}\\
			 \cline{7-8}
			 &      &    &    &   &      & \unet*     & \makecell{$\approx 6\%$ on average \\ ($\approx 7\%$ at 5-hr lead time)}\\
			 \cline{4-4}\cline{6-8}
			 
			 &      &    & \multirow{4}{*}{$\geq 10mm $} &   & \multirow{2}{*}{F1 score}   &  Persistence  & \makecell{$\approx 74\%$ on average \\ ($\approx 120\%$ at 5-hr lead time)}\\ 
			 \cline{7-8}
			 &      &    &    &   &       & \unet*      & \makecell{$\approx 30\%$ on average \\ ($\approx 84\%$ at 5-hr lead time)}\\  
			 \cline{6-8}
			 &      &    &    &   & \multirow{2}{*}{CSI score}      &  Persistence  & \makecell{$\approx 89\%$ on average \\ ($\approx 137\%$ at 5-hr lead time)}\\
			 \cline{7-8}
			 &      &    &    &   &       & \unet*         & \makecell{$\approx 32\%$ on average \\ ($\approx 96\%$ at 5-hr lead time)}\\
            \bottomrule
            \multicolumn{8}{l}{* \unet (\citep{agrawal2019machine}) trained using our loss function (e.g., \eqref{eq:loss:nowcasting}) without pre-training.}\\
        \end{tabular}}
    \label{tab:improve}
\end{table}
\black{Our deep-learning model assumes radar images collected from a fixed geographical area\black{;} thus\black{,} the training model cannot be applied to precipitation nowcasting in different areas. \cite{sonderby2020metnet} trained their model using patches covering different geographical areas with topographical data. As a result, the trained model can be used for multiple geographical areas and can even be applied to precipitation nowcasting in areas unseen during the training. It is worth applying this idea to our deep-learning model in the future.}

\black{\cite{ravuri2021skillful} pointed out that recent deep\black{-}learning-based precipitation nowcasting models, including \unet, produce blurry nowcasting images and thus may not include small-scale weather patterns. In order to produce a realistic and spatiotemporally consistent prediction, they suggested Deep Generative Models of Rainfall (DGMR) and trained adversarially with spatial and temporal discriminators, which discourage blurry predictions and temporally in-consistent predictions, respectively. Although DGMR still did not \black{show an improvement} over U-Net in terms of \black{the} CSI score, this idea also is worth applying to our study.}

\black{The aforementioned contributions are mostly orthogonal to our training strategies and \black{can therefore} be combined with ours. Our next step is to examine the impact of the combinations. In addition, we plan to evaluate the sensitivity of the performance of the deep-learning model developed according to the different synoptic systems, i.e., monsoon rainband\black{s}, mesoscale convective system\black{s}, extratropical and tropical cyclones. In particular, \black{w}e plan to deal with the sensitivity to \black{the} forecasting lead time for different heavy rainfall types with their performance's scientific interpretation. We expect to obtain ideas for improving \black{the} deep-learning model performance from a wider variety of perspectives through \black{a} quantitative evaluation.} 

\section{Conclusions} 
\label{sec:conclusion}
\black{In this work, we considered the problems of precipitation nowcasting and estimation from radar reflectivity images and} proposed effective training strategies for realizing the full potential of deep- learning-based approaches.
First, we proposed to pre-train the deep-learning model to predict radar reflectivity in the near \black{future before} fine-tuning it for a target task.
We noted that radar reflectivity, which is strongly correlated with precipitation rates, is typically available for a wide range of regions, while ground-truth precipitation is restricted only to weather stations.
Second, we proposed a novel loss function \black{(see Eq.~\eqref{eq:loss:nowcasting})} for mitigating the problems due to class imbalance\black{s} in precipitation data. 
Reducing the loss function requires accuracy not only for majority classes but also for minority classes, and thus minimizing the loss function prevents deep-learning models from ignoring minority classes.

While our training strategies are applicable to a wide range of deep-learning models, we demonstrated their effectiveness using a recent model based on U-Net after extending it for our tasks \black{and used} radar reflectivity and precipitation around South Korea collected over seven years.
Both our pre-training scheme and loss function significantly improved the accuracy of our deep-learning model.
For example, for precipitation nowcasting with five hours of lead time, our pre-training scheme improved \black{the} CSI and F1 scores for heavy rainfall (at least $10$ mm/hr) by up to $\mathbf{95.7\%}$ and up to $\mathbf{84.5\%}$, respectively, \black{compared to the fine-tuning only}, and using our loss function improved them up to $\mathbf{43.6\%}$ and up to $\mathbf{37.6\%}$, respectively, \black{compared to the conventional ones} \black{(see Tables~\ref{tab:pretrain} and \ref{tab:lossfn})}.
For precipitation estimation, which is formulated as a regression problem, the improvement due to pre-training in terms of MSE was up to $\mathbf{6.06\%}$ \black{(see Table~\ref{tab:resolution_est})}.
Lastly, we analyzed the sensitivity of our approach to \black{the} spatial resolution and \black{analyzed its performance for four cases of heavy rainfall.} \black{A h}igh resolution (i.e., $1$ km $\times$ $1$ km) was especially useful for being accurate for heavy rainfall, \black{though} \black{it showed limited ability when used to predict heavy rainfall} isolated in a small region 
\black{(see Tables~\ref{tab:resolution} and \ref{tab:rainfall})}.

\section*{Acknowledgements}
This work was supported by the Korea Meteorological Administration Research and Development Program under Grant KMI2020-01010 and \black{by a grant from the} Institute of Information
\& Communications Technology Planning \& Evaluation (IITP) funded by the Korea government (MSIT) (No. 2019-0-00075, Artificial Intelligence Graduate School Program (KAIST)).

\section*{Code availability}
The software presented in this work is named \textsc{DeepRaNE} (\textbf{Deep}-learning based \textbf{Ra}in \textbf{N}owcasting and \textbf{E}stimation) v1.0, 
and it is released under the GPL-3.0 license at \url{https://github.com/jihoonko/DeepRaNE}.
To run the software, Python 3.6.9 or later, NumPy 1.19 or later, and Pytorch 1.6.0 or later are required to be installed in the system.

\bibliographystyle{unsrtnat}
\bibliography{bibliography} 

\begin{thebibliography}{39}
\providecommand{\natexlab}[1]{#1}
\providecommand{\url}[1]{\texttt{#1}}
\expandafter\ifx\csname urlstyle\endcsname\relax
  \providecommand{\doi}[1]{doi: #1}\else
  \providecommand{\doi}{doi: \begingroup \urlstyle{rm}\Url}\fi

\bibitem[Shuman(1989)]{shuman1989history}
Frederick~G Shuman.
\newblock History of numerical weather prediction at the national
  meteorological center.
\newblock \emph{Weather and forecasting}, 4\penalty0 (3):\penalty0 286--296,
  1989.

\bibitem[Harper et~al.(2007)Harper, Uccellini, Kalnay, Carey, and
  Morone]{harper200750th}
Kristine Harper, Louis~W Uccellini, Eugenia Kalnay, Kenneth Carey, and Lauren
  Morone.
\newblock 50th anniversary of operational numerical weather prediction.
\newblock \emph{Bulletin of the American Meteorological Society}, 88\penalty0
  (5):\penalty0 639--650, 2007.

\bibitem[Marchuk(2012)]{marchuk2012numerical}
Gurii Marchuk.
\newblock \emph{Numerical methods in weather prediction}.
\newblock Elsevier, 2012.

\bibitem[Benjamin et~al.(2004)Benjamin, D{\'e}v{\'e}nyi, Weygandt, Brundage,
  Brown, Grell, Kim, Schwartz, Smirnova, Smith, et~al.]{benjamin2004hourly}
Stanley~G Benjamin, Dezs{\"o} D{\'e}v{\'e}nyi, Stephen~S Weygandt, Kevin~J
  Brundage, John~M Brown, Georg~A Grell, Dongsoo Kim, Barry~E Schwartz,
  Tatiana~G Smirnova, Tracy~Lorraine Smith, et~al.
\newblock An hourly assimilation--forecast cycle: The ruc.
\newblock \emph{Monthly Weather Review}, 132\penalty0 (2):\penalty0 495--518,
  2004.

\bibitem[Shrestha et~al.(2013)Shrestha, Robertson, Wang, Pagano, and
  Hapuarachchi]{shrestha2013evaluation}
DL~Shrestha, DE~Robertson, QJ~Wang, TC~Pagano, and HAP Hapuarachchi.
\newblock Evaluation of numerical weather prediction model precipitation
  forecasts for short-term streamflow forecasting purpose.
\newblock \emph{Hydrology and Earth System Sciences}, 17\penalty0 (5):\penalty0
  1913--1931, 2013.

\bibitem[Sun et~al.(2014)Sun, Xue, Wilson, Zawadzki, Ballard, Onvlee-Hooimeyer,
  Joe, Barker, Li, Golding, et~al.]{sun2014use}
Juanzhen Sun, Ming Xue, James~W Wilson, Isztar Zawadzki, Sue~P Ballard,
  Jeanette Onvlee-Hooimeyer, Paul Joe, Dale~M Barker, Ping-Wah Li, Brian
  Golding, et~al.
\newblock Use of nwp for nowcasting convective precipitation: Recent progress
  and challenges.
\newblock \emph{Bulletin of the American Meteorological Society}, 95\penalty0
  (3):\penalty0 409--426, 2014.

\bibitem[Wang et~al.(2016)Wang, Yang, Wang, and Liu]{wang2016quantitative}
Gaili Wang, Ji~Yang, Dan Wang, and Liping Liu.
\newblock A quantitative comparison of precipitation forecasts between the
  storm-scale numerical weather prediction model and auto-nowcast system in
  jiangsu, china.
\newblock \emph{Atmospheric Research}, 181:\penalty0 1--11, 2016.

\bibitem[Agrawal et~al.(2019)Agrawal, Barrington, Bromberg, Burge, Gazen, and
  Hickey]{agrawal2019machine}
Shreya Agrawal, Luke Barrington, Carla Bromberg, John Burge, Cenk Gazen, and
  Jason Hickey.
\newblock Machine learning for precipitation nowcasting from radar images.
\newblock \emph{arXiv preprint arXiv:1912.12132}, 2019.

\bibitem[S{\o}nderby et~al.(2020)S{\o}nderby, Espeholt, Heek, Dehghani, Oliver,
  Salimans, Agrawal, Hickey, and Kalchbrenner]{sonderby2020metnet}
Casper~Kaae S{\o}nderby, Lasse Espeholt, Jonathan Heek, Mostafa Dehghani,
  Avital Oliver, Tim Salimans, Shreya Agrawal, Jason Hickey, and Nal
  Kalchbrenner.
\newblock Metnet: A neural weather model for precipitation forecasting.
\newblock \emph{arXiv preprint arXiv:2003.12140}, 2020.

\bibitem[Ravuri et~al.(2021)Ravuri, Lenc, Willson, Kangin, Lam, Mirowski,
  Fitzsimons, Athanassiadou, Kashem, Madge, et~al.]{ravuri2021skillful}
Suman Ravuri, Karel Lenc, Matthew Willson, Dmitry Kangin, Remi Lam, Piotr
  Mirowski, Megan Fitzsimons, Maria Athanassiadou, Sheleem Kashem, Sam Madge,
  et~al.
\newblock Skillful precipitation nowcasting using deep generative models of
  radar.
\newblock \emph{arXiv preprint arXiv:2104.00954}, 2021.

\bibitem[Bechini and Chandrasekar(2017)]{bechini2017enhanced}
Renzo Bechini and V~Chandrasekar.
\newblock An enhanced optical flow technique for radar nowcasting of
  precipitation and winds.
\newblock \emph{Journal of Atmospheric and Oceanic Technology}, 34\penalty0
  (12):\penalty0 2637--2658, 2017.

\bibitem[Bowler et~al.(2004)Bowler, Pierce, and Seed]{bowler2004development}
Neill~EH Bowler, Clive~E Pierce, and Alan Seed.
\newblock Development of a precipitation nowcasting algorithm based upon
  optical flow techniques.
\newblock \emph{Journal of Hydrology}, 288\penalty0 (1-2):\penalty0 74--91,
  2004.

\bibitem[Seed et~al.(2013)Seed, Pierce, and Norman]{seed2013formulation}
Alan~W Seed, Clive~E Pierce, and Katie Norman.
\newblock Formulation and evaluation of a scale decomposition-based stochastic
  precipitation nowcast scheme.
\newblock \emph{Water Resources Research}, 49\penalty0 (10):\penalty0
  6624--6641, 2013.

\bibitem[Hwang et~al.(2015)Hwang, Clark, Lakshmanan, and
  Koch]{hwang2015improved}
Yunsung Hwang, Adam~J Clark, Valliappa Lakshmanan, and Steven~E Koch.
\newblock Improved nowcasts by blending extrapolation and model forecasts.
\newblock \emph{Weather and Forecasting}, 30\penalty0 (5):\penalty0 1201--1217,
  2015.

\bibitem[Shi et~al.(2015)Shi, Chen, Wang, Yeung, Wong, and
  Woo]{shi2015convolutional}
Xingjian Shi, Zhourong Chen, Hao Wang, Dit-Yan Yeung, Wai-Kin Wong, and
  Wang-chun Woo.
\newblock Convolutional lstm network: A machine learning approach for
  precipitation nowcasting.
\newblock In \emph{Advances in Neural Information Processing Systems (NIPS)},
  2015.

\bibitem[Shi et~al.(2017)Shi, Gao, Lausen, Wang, Yeung, Wong, and
  Woo]{shi2017deep}
Xingjian Shi, Zhihan Gao, Leonard Lausen, Hao Wang, Dit-Yan Yeung, Wai-Kin
  Wong, and Wang-chun Woo.
\newblock Deep learning for precipitation nowcasting: A benchmark and a new
  model.
\newblock In \emph{Advances in Neural Information Processing Systems (NIPS)},
  2017.

\bibitem[Ho et~al.(2019)Ho, Kalchbrenner, Weissenborn, and
  Salimans]{ho2019axial}
Jonathan Ho, Nal Kalchbrenner, Dirk Weissenborn, and Tim Salimans.
\newblock Axial attention in multidimensional transformers.
\newblock \emph{arXiv preprint arXiv:1912.12180}, 2019.

\bibitem[Ronneberger et~al.(2015)Ronneberger, Fischer, and
  Brox]{ronneberger2015u}
Olaf Ronneberger, Philipp Fischer, and Thomas Brox.
\newblock U-net: Convolutional networks for biomedical image segmentation.
\newblock In \emph{International Conference on Medical image computing and
  computer-assisted intervention}, pages 234--241, 2015.

\bibitem[Pulkkinen et~al.(2019)Pulkkinen, Nerini, P{\'e}rez~Hortal,
  Velasco-Forero, Seed, Germann, and Foresti]{pulkkinen2019pysteps}
Seppo Pulkkinen, Daniele Nerini, Andr{\'e}s~A P{\'e}rez~Hortal, Carlos
  Velasco-Forero, Alan Seed, Urs Germann, and Loris Foresti.
\newblock Pysteps: an open-source python library for probabilistic
  precipitation nowcasting (v1. 0).
\newblock \emph{Geoscientific Model Development}, 12\penalty0 (10):\penalty0
  4185--4219, 2019.

\bibitem[Benjamin et~al.(2016)Benjamin, Weygandt, Brown, Hu, Alexander,
  Smirnova, Olson, James, Dowell, Grell, et~al.]{benjamin2016north}
Stanley~G Benjamin, Stephen~S Weygandt, John~M Brown, Ming Hu, Curtis~R
  Alexander, Tatiana~G Smirnova, Joseph~B Olson, Eric~P James, David~C Dowell,
  Georg~A Grell, et~al.
\newblock A north american hourly assimilation and model forecast cycle: The
  rapid refresh.
\newblock \emph{Monthly Weather Review}, 144\penalty0 (4):\penalty0 1669--1694,
  2016.

\bibitem[Lebedev et~al.(2019)Lebedev, Ivashkin, Rudenko, Ganshin, Molchanov,
  Ovcharenko, Grokhovetskiy, Bushmarinov, and
  Solomentsev]{lebedev2019precipitation}
Vadim Lebedev, Vladimir Ivashkin, Irina Rudenko, Alexander Ganshin, Alexander
  Molchanov, Sergey Ovcharenko, Ruslan Grokhovetskiy, Ivan Bushmarinov, and
  Dmitry Solomentsev.
\newblock Precipitation nowcasting with satellite imagery.
\newblock In \emph{ACM SIGKDD International Conference on Knowledge Discovery
  \& Data Mining (KDD)}, 2019.

\bibitem[Ayzel et~al.(2020)Ayzel, Scheffer, and Heistermann]{ayzel2020rainnet}
Georgy Ayzel, Tobias Scheffer, and Maik Heistermann.
\newblock Rainnet v1. 0: a convolutional neural network for radar-based
  precipitation nowcasting.
\newblock \emph{Geoscientific Model Development}, 13\penalty0 (6):\penalty0
  2631--2644, 2020.

\bibitem[Japkowicz and Stephen(2002)]{japkowicz2002class}
Nathalie Japkowicz and Shaju Stephen.
\newblock The class imbalance problem: A systematic study.
\newblock \emph{Intelligent data analysis}, 6\penalty0 (5):\penalty0 429--449,
  2002.

\bibitem[Lin et~al.(2017)Lin, Goyal, Girshick, He, and
  Doll{\'a}r]{lin2017focal}
Tsung-Yi Lin, Priya Goyal, Ross Girshick, Kaiming He, and Piotr Doll{\'a}r.
\newblock Focal loss for dense object detection.
\newblock In \emph{IEEE international conference on computer vision (ICCV)},
  2017.

\bibitem[Sudre et~al.(2017)Sudre, Li, Vercauteren, Ourselin, and
  Cardoso]{sudre2017generalised}
Carole~H Sudre, Wenqi Li, Tom Vercauteren, Sebastien Ourselin, and M~Jorge
  Cardoso.
\newblock Generalised dice overlap as a deep learning loss function for highly
  unbalanced segmentations.
\newblock In \emph{Deep learning in medical image analysis and multimodal
  learning for clinical decision support}, pages 240--248. 2017.

\bibitem[Haltuf(2018)]{kaggle}
Michal Haltuf.
\newblock Best loss function for f1-score metric, 2018.
\newblock URL
  \url{https://www.kaggle.com/rejpalcz/best-loss-function-for-f1-score-metric}.

\bibitem[Donahue et~al.(2014)Donahue, Jia, Vinyals, Hoffman, Zhang, Tzeng, and
  Darrell]{donahue2014decaf}
Jeff Donahue, Yangqing Jia, Oriol Vinyals, Judy Hoffman, Ning Zhang, Eric
  Tzeng, and Trevor Darrell.
\newblock Decaf: A deep convolutional activation feature for generic visual
  recognition.
\newblock In \emph{ICML}, 2014.

\bibitem[Girshick et~al.(2014)Girshick, Donahue, Darrell, and
  Malik]{Girshick_2014_CVPR}
Ross Girshick, Jeff Donahue, Trevor Darrell, and Jitendra Malik.
\newblock Rich feature hierarchies for accurate object detection and semantic
  segmentation.
\newblock In \emph{IEEE Conference on Computer Vision and Pattern Recognition
  (CVPR)}, 2014.

\bibitem[Doersch and Zisserman(2017)]{doersch2017multi}
Carl Doersch and Andrew Zisserman.
\newblock Multi-task self-supervised visual learning.
\newblock In \emph{ICCV}, 2017.

\bibitem[Xie and Richmond(2018)]{xie2018pre}
Yiting Xie and David Richmond.
\newblock Pre-training on grayscale imagenet improves medical image
  classification.
\newblock In \emph{European Conference on Computer Vision Workshops}, 2018.

\bibitem[Rasp and Thuerey(2021)]{rasp2021data}
Stephan Rasp and Nils Thuerey.
\newblock Data-driven medium-range weather prediction with a resnet pretrained
  on climate simulations: A new model for weatherbench.
\newblock \emph{Journal of Advances in Modeling Earth Systems}, 13\penalty0
  (2):\penalty0 e2020MS002405, 2021.

\bibitem[He et~al.(2016)He, Zhang, Ren, and Sun]{he2016deep}
Kaiming He, Xiangyu Zhang, Shaoqing Ren, and Jian Sun.
\newblock Deep residual learning for image recognition.
\newblock In \emph{IEEE conference on computer vision and pattern recognition},
  2016.

\bibitem[Eyring et~al.(2016)Eyring, Bony, Meehl, Senior, Stevens, Stouffer, and
  Taylor]{eyring2016overview}
Veronika Eyring, Sandrine Bony, Gerald~A Meehl, Catherine~A Senior, Bjorn
  Stevens, Ronald~J Stouffer, and Karl~E Taylor.
\newblock Overview of the coupled model intercomparison project phase 6 (cmip6)
  experimental design and organization.
\newblock \emph{Geoscientific Model Development}, 9\penalty0 (5):\penalty0
  1937--1958, 2016.

\bibitem[Hersbach et~al.(2020)Hersbach, Bell, Berrisford, Hirahara,
  Hor{\'a}nyi, Mu{\~n}oz-Sabater, Nicolas, Peubey, Radu, Schepers,
  et~al.]{hersbach2020era5}
Hans Hersbach, Bill Bell, Paul Berrisford, Shoji Hirahara, Andr{\'a}s
  Hor{\'a}nyi, Joaqu{\'\i}n Mu{\~n}oz-Sabater, Julien Nicolas, Carole Peubey,
  Raluca Radu, Dinand Schepers, et~al.
\newblock The era5 global reanalysis.
\newblock \emph{Quarterly Journal of the Royal Meteorological Society},
  146\penalty0 (730):\penalty0 1999--2049, 2020.

\bibitem[Rubner et~al.(2000)Rubner, Tomasi, and Guibas]{rubner2000earth}
Yossi Rubner, Carlo Tomasi, and Leonidas~J Guibas.
\newblock The earth mover's distance as a metric for image retrieval.
\newblock \emph{International Journal of Computer Vision (IJCV)}, 40\penalty0
  (2):\penalty0 99--121, 2000.

\bibitem[Marshall(1948)]{marshall1948distribution}
John~S Marshall.
\newblock The distribution of raindrops with size.
\newblock \emph{J. meteor.}, 5:\penalty0 165--166, 1948.

\bibitem[Kingma and Ba(2015)]{kingma2014adam}
Diederik~P Kingma and Jimmy Ba.
\newblock Adam: A method for stochastic optimization.
\newblock In \emph{International Conference on Learning Representations
  (ICLR)}, 2015.

\bibitem[Ioffe and Szegedy(2015)]{ioffe2015batch}
Sergey Ioffe and Christian Szegedy.
\newblock Batch normalization: Accelerating deep network training by reducing
  internal covariate shift.
\newblock In \emph{ICML}, 2015.

\bibitem[Nair and Hinton(2010)]{nair2010rectified}
Vinod Nair and Geoffrey~E Hinton.
\newblock Rectified linear units improve restricted boltzmann machines.
\newblock In \emph{ICML}, 2010.

\end{thebibliography}
\newpage

\appendix
\numberwithin{equation}{section}
\renewcommand{\theequation}{A.\arabic{equation}}

\section{A detailed description of the model architecture and the loss functions}

In this section, we describe the model architecture in Figure \ref{fig:model} and the loss functions used in Section \ref{sec:experiments:nowcasting}.

\subsection{Model architecture}
\label{sec:app:model}
\unet \citep{ronneberger2015u} consists of a contracting path and an expansive path.
We used two convolutional layers in the contracting path with a receptive field size of 3$\times$3 (depicted with yellow arrows in Figure \ref{fig:model}), to process the initial input feature maps and the feature maps after each downsampling \black{step}. When input feature maps $X \in \mathbb{R}^{c_{in} \times h \times w}$ are given, through the convolutional layer, each element of the output feature maps $Y \in \mathbb{R}^{c_{out} \times (h-2) \times (w-2)}$ is computed by Eq.~\eqref{eq:unet:conv}.
\begin{align}
    Y^{(c)}_{i,j} &= \sum_{c'=1}^{c_{in}} \sum_{i'=1}^{3} \sum_{j'=1}^{3} F^{c,c'}_{i', j'} X^{(c')}_{i+i'-1, j+j'-1} \qquad \forall i \in \{1, 2, \ldots, h-2\}, j \in \{1, 2, \ldots, w-2\}, c \in \{1, 2, \ldots, c_{out}\},
    \label{eq:unet:conv}
\end{align}
where the entries of $F^{c,c'} \in \mathbb{R}^{3 \times 3}$ are the learnable parameters of the layer. For the convolutional layer \black{immediately} right after the downsampling \black{step}, we set $c_{out}$ to twice \black{the value} of $c_{in}$. Batch normalization (BN) \citep{ioffe2015batch} and ReLU \citep{nair2010rectified} activation are additionally used \black{directly} after the convolution operation \black{to stabilize} learning and \black{to ensure that} the model \black{is} non-linear.

For downsampling, a max-pooling operation with a 2$\times$2 filter (depicted with blue arrows in Figure \ref{fig:model}) is used. The operation \black{initially} considers each input feature map as a block matrix\black{,} where all blocks are 2$\times$2 \black{in size,} and then finds the maximum value in each block.
When input feature maps $X \in \mathbb{R}^{c_{in} \times h \times w}$ are given, after the max pooling operation, each element of the output feature maps $Y \in \mathbb{R}^{c_{in} \times (h/2) \times (w/2)}$ is computed by Eq.~\eqref{eq:unet:maxpool}.
\begin{align}
     Y^{(c)}_{i,j} &= \max(X^{(c)}_{2i-1, 2j-1}, X^{(c)}_{2i-1, 2j}, X^{(c)}_{2i, 2j-1}, X^{(c)}_{2i, 2j}) \quad \forall i \in \{1, 2, \ldots, h/2\}.j \in \{1, 2, \ldots, w/2\}, c \in \{1, 2, \ldots, c_{in}\}
     \label{eq:unet:maxpool}
\end{align}

\black{As an example step of the contracting path, consider the feature maps at the beginning. After the first two convolution operations, the feature maps are of \black{a} dimension \black{of }$32$ (channels) $\times$ $1464$ (height) $\times$ $1464$ (width). By applying 2$\times$2 max pooling, the width and height of the feature map are reduced to half, and the number of channels is unchanged. Then, the pooled feature maps are transformed again by the two convolutional layers \black{directly} after max-pooling. The\black{se} operations double the number of channels and decrease both the width and height by $4$. Thus, the feature maps right before the second max-pooling are of \black{a} dimension \black{of} $64 \times 728 \times 728$. Max-pooling operations are performed seven times to encode the inputs \black{compactly}.}

\black{In the expansive path, the following process is repeated seven times: (a) The contracted feature maps are up-sampled using \black{an} up-convolution operation (depicted \black{by the} red arrows in Figure \ref{fig:model}). (b) Then, they are concatenated with intermediate feature maps from the contracting path (depicted by the gray arrows in Figure \ref{fig:model}). (c) \black{Subsequently}, two convolutional layers with a receptive field size of $3\times3$ (depicted \black{by the} yellow arrows in Figure \ref{fig:model}) are applied.}

\black{When upsampling the feature maps, \black{an} up-convolution operation, which is also called transposed convolution, is applied. Unlike the convolution operation, the up-convolution operation increases the height and width of the feature maps. The height and width are doubled in our setting, while the number of channels is reduced \black{by} half. After the $i$-th up-convolution operation, the output and the intermediate feature maps \black{immediately} before the $(8-i)$-th max-pooling from the contracting path are concatenated. \black{Because} the width and height of the latter are greater those of the former, the latter is cropped from the center of the feature map to match the size. Then, the two feature maps are concatenated and used as the input of the convolutional layer right after the concatenation \black{step}. Unlike in the contracting path, $c_{out}$ is set to half of $c_{in}$.}

\black{As an example step for the expansive path, consider the feature maps before the last up-convolution. The input of the last up-convolution is of \black{a dimension of} $64$ (channels) $\times$ $356$ (height) $\times$ $356$ (width), and the transformed feature maps after the operation is of \black{a dimension of} $32 \times 712 \times 712$. when cropping the inputs of the contracting path from the center and concatenating it with the upsampled feature maps, the number of output channels is doubled, while the width and height do not change. After the two convolution operations, the number of output channels becomes $32$ again, and the width and height of the output decrease by $4$.} 

\black{To compute the final output, an additional 3$\times$3 convolutional layer is added at the end of the path, and the number of channels of the output depends on the task. The final output dimension of the model is 706$\times$706, and \black{the} $(i,j)$-th pixel of the model's output corresponds to \black{the} $(i+381,j+381)$-th pixel of the input of the model. 
That is, the set of regions of interest is \black{a} patch of \black{a} dimension of $706$ $\times$ $706$ at the center of the radar images.}

\subsection{Loss functions}
\label{sec:app:loss}
\blue{The cross-entropy (CE) loss is a widely-used loss function in classification tasks.
For each instance, we measure the cross entropy between the predicted probability distribution $q$ over classes and the ground-truth distribution $p$, which is defined as: 
\begin{equation}
    CE(p, q) = -\sum_{x \in \mathcal{X}} p(x) \cdot \log q(x),
\end{equation}
where $\mathcal{X} = \{ \norain, \lightrain, \heavyrain \}$ in our case. 
\black{Given that} the probability mass function (PMF) of the ground-truth distribution is always $1$ for the ground-truth class and $0$ otherwise, the cross entropy is expressed as \begin{equation}
CE(p,q) = -\log(q(x^*_p)),
\end{equation}
where $x^*_p\in \mathcal{X}$ is the ground-truth class where $p(x^*_p)=1$. The training loss is defined as the sum of the cross entropy for all the instances in the training set.}

\black{The focal loss (FL) \citep{lin2017focal} adjusts the importance of each instance by multiplying $(1-q(x^*_p))^{\gamma}$ \black{by the} cross entropy, as follows:
\begin{equation}
    FL(p, q) = -(1-q(x^*_p))^{\gamma} \cdot CE(p, q) = -(1-q(x^*_p))^{\gamma} \cdot \log(q(x^*_p)),
\end{equation}
where $\gamma \geq 0$ is a hyperparameter. As $\gamma$ increases, the importance of ill-classified examples (i.e., where $q(x^*_p) \approx 0$) relative to that of well-classified examples (i.e., where $q(x^*_p) \approx 1$) increases.}

\blue{\section{Confusion matrix}
\label{sec:app:confusion}
In Tables~\ref{tab:confusion_pretrain_new}-\ref{tab:confusion_ce},
we provide \black{a} confusion matrix at each lead time in four settings: (a) using our loss function (Eq.~\eqref{eq:loss:nowcasting}) with pre-training, (b) using our loss function without pre-training, (c) using the focal loss \citep{lin2017focal} with pre-training, and (d) using the cross entropy loss with pre-training.}

\begin{table}[h]
    \centering
    \caption{
    \blue{Confusion matrices when our loss function (i.e., Eq.~\eqref{eq:loss:nowcasting}) was used with pre-training. The ratios of overestimation (O.R.)  and underestimation (U.R.) averaged over all lead times are 5.41\% and 2.86\%, respectively.}
    }
    \begin{subtable}[h]{0.32\textwidth}
    \centering
        \scalebox{0.7}{
        \begin{tabular}{c|c|c|c}
            \toprule
            \backslashbox{Actual\kern-1.2em}{\kern-1.2em Predicted} & \norain & \lightrain & \heavyrain \\
            \hline\hline\rule{0pt}{2.5ex}
            \norain	&	1842535 	&	58886 	&	1229  \\
            \lightrain	&	28095 	&	110118 	&	5970  \\
            \heavyrain	&	203 	&	10174 	&	11254  \\
            \bottomrule
        \end{tabular}
        }
        \caption{Lead time = $1$ hour \newline \blue{(O.R. = 3.19\%, U.R. = 1.86\%)}}
    \end{subtable}
    \hspace{0mm}
    \begin{subtable}[h]{0.32\textwidth}
    \centering
        \scalebox{0.7}{
        \begin{tabular}{c|c|c|c}
            \toprule
            \backslashbox{Actual\kern-1.2em}{\kern-1.2em Predicted} & \norain & \lightrain & \heavyrain \\
            \hline\hline\rule{0pt}{2.5ex}
            \norain	&	1816885 	&	82091 	&	3674  \\
            \lightrain	&	38811 	&	88713 	&	16659  \\
            \heavyrain	&	1109 	&	8758 	&	11764  \\
            \bottomrule
        \end{tabular}
        }
        \caption{Lead time = $2$ hours \newline \blue{(O.R. = 4.95\%, U.R. = 2.35\%)}}
    \end{subtable}
    \hspace{0mm}
    \begin{subtable}[h]{0.32\textwidth}
    \centering
        \scalebox{0.7}{
        \begin{tabular}{c|c|c|c}
            \toprule
            \backslashbox{Actual\kern-1.2em}{\kern-1.2em Predicted} & \norain & \lightrain & \heavyrain \\
            \hline\hline\rule{0pt}{2.5ex}
            \norain	&	1800305 	&	94859 	&	7496  \\
            \lightrain	&	43167 	&	78158 	&	22858  \\
            \heavyrain	&	2222 	&	8470 	&	10939  \\
            \bottomrule
        \end{tabular}
        }
        \caption{Lead time = $3$ hours \newline\blue{(O.R. = 6.05\%, U.R. = 2.60\%)}}
    \end{subtable}
    \begin{subtable}[h]{0.32\textwidth}
    \centering
        \scalebox{0.7}{
        \begin{tabular}{c|c|c|c}
            \toprule
            \backslashbox{Actual\kern-1.2em}{\kern-1.2em Predicted} & \norain & \lightrain & \heavyrain \\
            \hline\hline\rule{0pt}{2.5ex}
            \norain	&	1798917 	&	93022 	&	10711  \\
            \lightrain	&	51344 	&	70094 	&	22745  \\
            \heavyrain	&	3688 	&	8567 	&	9376  \\
            \bottomrule
        \end{tabular}
        }
        \caption{Lead time = $4$ hours \newline\blue{(O.R. = 6.11\%, U.R. = 3.07\%)}}
    \end{subtable}
    \hspace{0mm}
    \begin{subtable}[h]{0.32\textwidth}
    \centering
        \scalebox{0.7}{
        \begin{tabular}{c|c|c|c}
            \toprule
            \backslashbox{Actual\kern-1.2em}{\kern-1.2em Predicted} & \norain & \lightrain & \heavyrain \\
            \hline\hline\rule{0pt}{2.5ex}
            \norain	&	1802681 	&	90368 	&	9611  \\
            \lightrain	&	59751 	&	68603 	&	15829  \\
            \heavyrain	&	4812 	&	10454 	&	6365  \\
            \bottomrule
        \end{tabular}
        }
        \caption{Lead time = $5$ hours \newline\blue{(O.R. = 5.60\%, U.R. = 3.63\%)}}
    \end{subtable}
    \hspace{0mm}
    \begin{subtable}[h]{0.32\textwidth}
    \centering
        \scalebox{0.7}{
        \begin{tabular}{c|c|c|c}
            \toprule
            \backslashbox{Actual\kern-1.2em}{\kern-1.2em Predicted} & \norain & \lightrain & \heavyrain \\
            \hline\hline\rule{0pt}{2.5ex}
            \norain	&	1783689                      & 105693                     & 13268  \\
            \lightrain	&	59689                        & 67599                      & 16895  \\
            \heavyrain	&	5411                         & 10236                      & 5984  \\
            \bottomrule
        \end{tabular}
        }
        \caption{Lead time = $6$ hours \newline\blue{(O.R. = 6.57\%, U.R. = 3.64\%)}}
    \end{subtable}
    \label{tab:confusion_pretrain_new}
\end{table}

\begin{table}[h]
    \centering
    \caption{
    \blue{Confusion matrices when our loss function (i.e., Eq.~\eqref{eq:loss:nowcasting}) was used without pre-training. The ratios of overestimation (O.R.)  and underestimation (U.R.) averaged over all lead times are 3.45\% and  3.77\%, respectively.}
    }
    \begin{subtable}[h]{0.32\textwidth}
    \centering
        \scalebox{0.7}{
        \begin{tabular}{c|c|c|c}
            \toprule
            \backslashbox{Actual\kern-1.2em}{\kern-1.2em Predicted} & \norain & \lightrain & \heavyrain \\
            \hline\hline\rule{0pt}{2.5ex}
            \norain	& 1850542 & 50632 & 1486  \\
            \lightrain	& 34430   & 97668 & 12085  \\
            \heavyrain	& 679     & 7704  & 13248  \\
            \bottomrule
        \end{tabular}
        }
        \caption{Lead time = $1$ hour \newline\blue{(O.R. = 3.10\%, U.R. = 2.07\%)}}
    \end{subtable}
    \hspace{0mm}
    \begin{subtable}[h]{0.32\textwidth}
    \centering
        \scalebox{0.7}{
        \begin{tabular}{c|c|c|c}
            \toprule
            \backslashbox{Actual\kern-1.2em}{\kern-1.2em Predicted} & \norain & \lightrain & \heavyrain \\
            \hline\hline\rule{0pt}{2.5ex}
            \norain	& 1833979 & 64311 & 4370  \\
            \lightrain	& 47034   & 76400 & 20749  \\
            \heavyrain	& 2300    & 8793  & 10538  \\
            \bottomrule
        \end{tabular}
        }
        \caption{Lead time = $2$ hours \newline\blue{(O.R. = 4.32\%, U.R. = 2.81\%)}}
    \end{subtable}
    \hspace{0mm}
    \begin{subtable}[h]{0.32\textwidth}
    \centering
        \scalebox{0.7}{
        \begin{tabular}{c|c|c|c}
            \toprule
            \backslashbox{Actual\kern-1.2em}{\kern-1.2em Predicted} & \norain & \lightrain & \heavyrain \\
            \hline\hline\rule{0pt}{2.5ex}
            \norain	&	1823363 & 75620 & 3677  \\
            \lightrain	&	54560   & 76389 & 13234  \\
            \heavyrain	&	3904    & 10819 & 6908 \\
            \bottomrule
        \end{tabular}
        }
        \caption{Lead time = $3$ hours \newline\blue{(O.R. = 4.47\%, U.R. = 3.35\%)}}
    \end{subtable}
    \begin{subtable}[h]{0.32\textwidth}
    \centering
        \scalebox{0.7}{
        \begin{tabular}{c|c|c|c}
            \toprule
            \backslashbox{Actual\kern-1.2em}{\kern-1.2em Predicted} & \norain & \lightrain & \heavyrain \\
            \hline\hline\rule{0pt}{2.5ex}
            \norain	&	1820495 & 77787 & 4378  \\
            \lightrain	&	59577   & 73718 & 10888  \\
            \heavyrain	&	5203    & 11386 & 5042  \\
            \bottomrule
        \end{tabular}
        }
        \caption{Lead time = $4$ hours \newline\blue{(O.R. = 4.50\%, U.R. = 3.68\%)}}
    \end{subtable}
    \hspace{0mm}
    \begin{subtable}[h]{0.32\textwidth}
    \centering
        \scalebox{0.7}{
        \begin{tabular}{c|c|c|c}
            \toprule
            \backslashbox{Actual\kern-1.2em}{\kern-1.2em Predicted} & \norain & \lightrain & \heavyrain \\
            \hline\hline\rule{0pt}{2.5ex}
            \norain	&	1845752 & 54771 & 2137  \\
            \lightrain	&	78372   & 62195 & 3616  \\
            \heavyrain	&	8191    & 11559 & 1881  \\
            \bottomrule
        \end{tabular}
        }
        \caption{Lead time = $5$ hours \newline\blue{(O.R. = 2.93\%, U.R. = 4.74\%)}}
    \end{subtable}
    \hspace{0mm}
    \begin{subtable}[h]{0.32\textwidth}
    \centering
        \scalebox{0.7}{
        \begin{tabular}{c|c|c|c}
            \toprule
            \backslashbox{Actual\kern-1.2em}{\kern-1.2em Predicted} & \norain & \lightrain & \heavyrain \\
            \hline\hline\rule{0pt}{2.5ex}
            \norain	&	1873770 & 28782 & 108  \\
            \lightrain	&	102457  & 41591 & 135  \\
            \heavyrain	&	12544   & 8985  & 102  \\
            \bottomrule
        \end{tabular}
        }
        \caption{Lead time = $6$ hours \newline\blue{(O.R. = 1.40\%, U.R. = 5.99\%)}}
    \end{subtable}
    \label{tab:confusion_nopretrain_new}
\end{table}

\begin{table}[t]
    \centering
    \caption{
    \blue{Confusion matrices when the focal loss \cite{lin2017focal} was used with pre-training. The ratios of overestimation (O.R.)  and underestimation (U.R.) averaged over all lead times are 3.09\% and  3.70\%, respectively.}}
    \begin{subtable}[h]{0.32\textwidth}
    \centering
        \scalebox{0.7}{
        \begin{tabular}{c|c|c|c}
            \toprule
            \backslashbox{Actual\kern-1.2em}{\kern-1.2em Predicted} & \norain & \lightrain & \heavyrain \\
            \hline\hline\rule{0pt}{2.5ex}
            \norain	&	1865533 & 36166  & 961  \\
            \lightrain	&	35648   & 102882 & 5653  \\
            \heavyrain	&	423     & 9854   & 11354  \\
            \bottomrule
        \end{tabular}
        }
        \caption{Lead time = $1$ hour \newline\blue{(O.R. = 2.07\%, U.R. = 2.22\%)}}
    \end{subtable}
    \hspace{0mm}
    \begin{subtable}[h]{0.32\textwidth}
    \centering
        \scalebox{0.7}{
        \begin{tabular}{c|c|c|c}
            \toprule
            \backslashbox{Actual\kern-1.2em}{\kern-1.2em Predicted} & \norain & \lightrain & \heavyrain \\
            \hline\hline\rule{0pt}{2.5ex}
            \norain	&	1857011 & 44217 & 1432  \\
            \lightrain	&	51107   & 82889 & 10187  \\
            \heavyrain	&	2400    & 10737 & 8494  \\
            \bottomrule
        \end{tabular}
        }
        \caption{Lead time = $2$ hours \newline\blue{(O.R. = 2.70\%, U.R. = 3.11\%)}}
    \end{subtable}
    \hspace{0mm}
    \begin{subtable}[h]{0.32\textwidth}
    \centering
        \scalebox{0.7}{
        \begin{tabular}{c|c|c|c}
            \toprule
            \backslashbox{Actual\kern-1.2em}{\kern-1.2em Predicted} & \norain & \lightrain & \heavyrain \\
            \hline\hline\rule{0pt}{2.5ex}
            \norain	&	1852840 & 47612 & 2208  \\
            \lightrain	&	63055   & 71271 & 9857  \\
            \heavyrain	&	4336    & 10979 & 6316  \\
            \bottomrule
        \end{tabular}
        }
        \caption{Lead time = $3$ hours \newline\blue{(O.R. = 2.89\%, U.R. = 3.79\%)}}
    \end{subtable}
    \begin{subtable}[h]{0.32\textwidth}
    \centering
        \scalebox{0.7}{
        \begin{tabular}{c|c|c|c}
            \toprule
            \backslashbox{Actual\kern-1.2em}{\kern-1.2em Predicted} & \norain & \lightrain & \heavyrain \\
            \hline\hline\rule{0pt}{2.5ex}
            \norain	&	1845446 & 54432 & 2782  \\
            \lightrain	&	69601   & 66377 & 8205  \\
            \heavyrain	&	6134    & 11260 & 4237  \\
            \bottomrule
        \end{tabular}
        }
        \caption{Lead time = $4$ hours \newline\blue{(O.R. = 3.16\%, U.R. = 4.21\%)}}
    \end{subtable}
    \hspace{0mm}
    \begin{subtable}[h]{0.32\textwidth}
    \centering
        \scalebox{0.7}{
        \begin{tabular}{c|c|c|c}
            \toprule
            \backslashbox{Actual\kern-1.2em}{\kern-1.2em Predicted} & \norain & \lightrain & \heavyrain \\
            \hline\hline\rule{0pt}{2.5ex}
            \norain	&	1838026 & 61362 & 3272  \\
            \lightrain	&	73949   & 63312 & 6922  \\
            \heavyrain	&	7124    & 11500 & 3007  \\
            \bottomrule
        \end{tabular}
        }
        \caption{Lead time = $5$ hours \newline\blue{(O.R. = 3.46\%, U.R. = 4.48\%)}}
    \end{subtable}
    \hspace{0mm}
    \begin{subtable}[h]{0.32\textwidth}
    \centering
        \scalebox{0.7}{
        \begin{tabular}{c|c|c|c}
            \toprule
            \backslashbox{Actual\kern-1.2em}{\kern-1.2em Predicted} & \norain & \lightrain & \heavyrain \\
            \hline\hline\rule{0pt}{2.5ex}
            \norain	&	1821425 & 77036 & 4199  \\
            \lightrain	&	72642   & 64647 & 6894  \\
            \heavyrain	&	7612    & 11360 & 2659  \\
            \bottomrule
        \end{tabular}
        }
        \caption{Lead time = $6$ hours \newline\blue{(O.R. = 4.26\%, U.R. = 4.43\%)}}
    \end{subtable}
    \label{tab:confusion_focal}
\end{table}

\begin{table}[t]
    \centering
    \caption{
     \blue{Confusion matrices when the cross entropy loss was used with pre-training. The ratios of overestimation (O.R.)  and underestimation (U.R.) averaged over all lead times are 3.03\% and  3.74\%, respectively.}
    }
    \begin{subtable}[h]{0.32\textwidth}
    \centering
        \scalebox{0.7}{
        \begin{tabular}{c|c|c|c}
            \toprule
            \backslashbox{Actual\kern-1.2em}{\kern-1.2em Predicted} & \norain & \lightrain & \heavyrain \\
            \hline\hline\rule{0pt}{2.5ex}
            \norain	&	1854984 & 46015  & 1661  \\
            \lightrain	&	28483   & 100883 & 14817  \\
            \heavyrain	&	407     & 5129   & 16095  \\
            \bottomrule
        \end{tabular}
        }
        \caption{Lead time = $1$ hour \newline\blue{(O.R. = 3.02\%, U.R. = 1.64\%)}}
    \end{subtable}
    \hspace{0mm}
    \begin{subtable}[h]{0.32\textwidth}
    \centering
        \scalebox{0.7}{
        \begin{tabular}{c|c|c|c}
            \toprule
            \backslashbox{Actual\kern-1.2em}{\kern-1.2em Predicted} & \norain & \lightrain & \heavyrain \\
            \hline\hline\rule{0pt}{2.5ex}
            \norain	&	1850883 & 50180 & 1597  \\
            \lightrain	&	48484   & 85581 & 10118  \\
            \heavyrain	&	2436    & 10617 & 8578  \\
            \bottomrule
        \end{tabular}
        }
        \caption{Lead time = $2$ hours \newline\blue{(O.R. = 2.99\%, U.R. = 2.97\%)}}
    \end{subtable}
    \hspace{0mm}
    \begin{subtable}[h]{0.32\textwidth}
    \centering
        \scalebox{0.7}{
        \begin{tabular}{c|c|c|c}
            \toprule
            \backslashbox{Actual\kern-1.2em}{\kern-1.2em Predicted} & \norain & \lightrain & \heavyrain \\
            \hline\hline\rule{0pt}{2.5ex}
            \norain	&	1850154 & 51179 & 1327  \\
            \lightrain	&	63431   & 74555 & 6197  \\
            \heavyrain	&	4753    & 12493 & 4385  \\
            \bottomrule
        \end{tabular}
        }
        \caption{Lead time = $3$ hours \newline\blue{(O.R. = 2.84\%, U.R. = 3.90\%)}}
    \end{subtable}
    \hspace{0mm}
    \begin{subtable}[h]{0.32\textwidth}
    \centering
        \scalebox{0.7}{
        \begin{tabular}{c|c|c|c}
            \toprule
            \backslashbox{Actual\kern-1.2em}{\kern-1.2em Predicted} & \norain & \lightrain & \heavyrain \\
            \hline\hline\rule{0pt}{2.5ex}
            \norain	&	1845129 & 56172 & 1359  \\
            \lightrain	&	70671   & 68935 & 4577  \\
            \heavyrain	&	6340    & 12684 & 2607  \\
            \bottomrule
        \end{tabular}
        }
        \caption{Lead time = $4$ hours \newline\blue{(O.R. = 3.00\%, U.R. = 4.34\%)}}
    \end{subtable}
    \hspace{0mm}
    \begin{subtable}[h]{0.32\textwidth}
    \centering
        \scalebox{0.7}{
        \begin{tabular}{c|c|c|c}
            \toprule
            \backslashbox{Actual\kern-1.2em}{\kern-1.2em Predicted} & \norain & \lightrain & \heavyrain \\
            \hline\hline\rule{0pt}{2.5ex}
            \norain	&	1839613 & 61581 & 1466  \\
            \lightrain	&	75864   & 65115 & 3204  \\
            \heavyrain	&	7634    & 12401 & 1596  \\
            \bottomrule
        \end{tabular}
        }
        \caption{Lead time = $5$ hours \newline\blue{(O.R. = 3.20\%, U.R. = 4.64\%)}}
    \end{subtable}
    \hspace{0mm}
    \begin{subtable}[h]{0.32\textwidth}
    \centering
        \scalebox{0.7}{
        \begin{tabular}{c|c|c|c}
            \toprule
            \backslashbox{Actual\kern-1.2em}{\kern-1.2em Predicted} & \norain & \lightrain & \heavyrain \\
            \hline\hline\rule{0pt}{2.5ex}
            \norain	&	1840242 & 61120 & 1298  \\
            \lightrain	&	82083   & 60217 & 1883  \\
            \heavyrain	&	9076    & 11774 & 781  \\
            \bottomrule
        \end{tabular}
        }
        \caption{Lead time = $6$ hours \newline\blue{(O.R. = 3.11\%, U.R. = 4.98\%)}}
    \end{subtable}
    \label{tab:confusion_ce}
\end{table}

\end{document}